\documentclass[10pt,twocolumn,letterpaper]{article}
\usepackage{iccv}
\usepackage{times}
\usepackage{epsfig}
\usepackage{graphicx}
\usepackage{amsmath}
\usepackage{amssymb}
\usepackage{multirow}
\usepackage{caption}
\usepackage{csquotes}
\usepackage{bm}
\usepackage{hhline}
 \iccvfinalcopy 

\def\iccvPaperID{10644} 

\ificcvfinal\fi

\newcommand{\wrd}[1]{{\small\texttt{#1}}}
\usepackage{iccv}
\usepackage{times}
\usepackage{epsfig}
\usepackage{graphicx}
\usepackage{amsmath}
\usepackage{amssymb}
\usepackage{subfigure}
 \usepackage{caption}
 \usepackage{multirow}
 \usepackage{hhline}

\usepackage[pagebackref=true,breaklinks=true,letterpaper=true,colorlinks,bookmarks=false]{hyperref}

\iccvfinalcopy 

\def\iccvPaperID{10644} 

\ificcvfinal\fi

\begin{document}

\title{Handwriting Transformers}
\author{Ankan Kumar Bhunia\textsuperscript{1} \hspace{.1cm} Salman Khan\textsuperscript{1,2} \hspace{.1cm} Hisham Cholakkal\textsuperscript{1} \hspace{.1cm} Rao Muhammad Anwer\textsuperscript{1}\\Fahad Shahbaz Khan\textsuperscript{1,3} \hspace{.1cm} 
Mubarak Shah\textsuperscript{4}\\
\textsuperscript{1}Mohamed bin Zayed University of AI, UAE \hspace{.1cm} \textsuperscript{2}Australian National University, Australia
\hspace{.1cm}  \\ \textsuperscript{3}Link{\"o}ping University, Sweden  \hspace{.1cm} \textsuperscript{4}University of Central Florida, USA\\
}
\maketitle

\maketitle
\ificcvfinal\thispagestyle{empty}\fi

\begin{abstract}
We propose a novel transformer-based styled handwritten text image generation approach, HWT, that strives to learn both style-content entanglement as well as global and local writing style patterns. The proposed HWT captures the long and short range  relationships within the style examples through a self-attention mechanism, thereby encoding both global and local style patterns. Further, the proposed transformer-based HWT comprises an encoder-decoder attention that enables style-content entanglement by gathering the style representation of each query character. To the best of our knowledge, we are the first to introduce a transformer-based generative network for styled handwritten text generation.
 
Our proposed HWT generates realistic styled handwritten text images and significantly outperforms the state-of-the-art demonstrated through extensive qualitative, quantitative and human-based evaluations. The proposed HWT can handle arbitrary length of text and any desired writing style in a few-shot setting. Further, our HWT generalizes well to the challenging scenario where both words and writing style are \textit{unseen} during training, generating realistic styled handwritten text images.

\end{abstract}

\section{Introduction}

\begin{figure}[t!]
  \includegraphics[width=\linewidth]{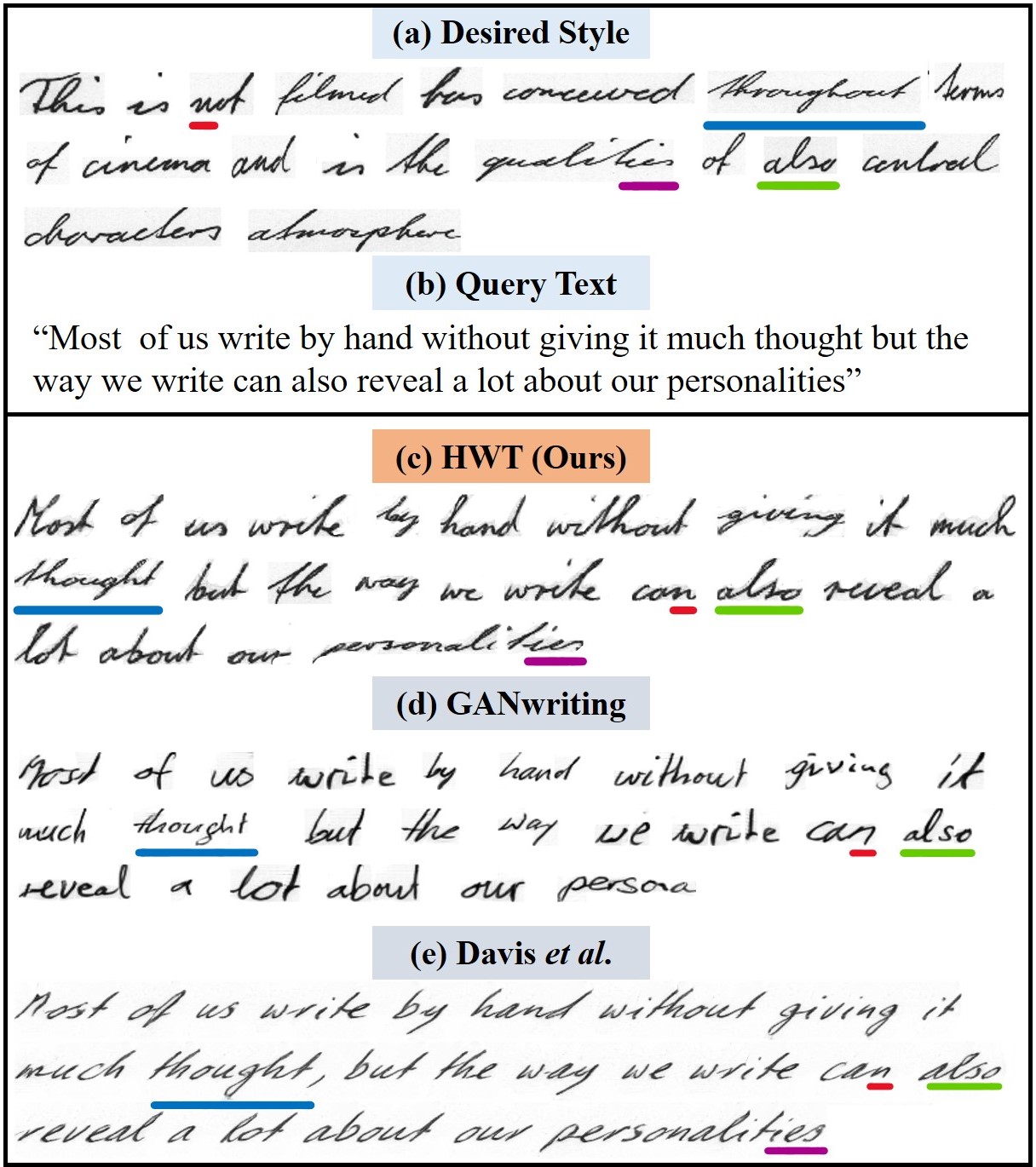}\vspace{-0.1cm}
      \caption{Comparison of HWT \textbf{(c)} with GANwriting~\cite{kang2020ganwriting} \textbf{(d)} and Davis~\etal~\cite{davis2020text} \textbf{(e)} in imitating the desired unseen writing style \textbf{(a)} for given query text \textbf{(b)}. While~\cite{kang2020ganwriting,davis2020text} capture global writing styles (\eg, slant), they struggle to imitate local style patterns (\eg, character style, ligatures). HWT (c) imitates both global and local styles, leading to a more realistic styled handwritten text image generation. For instance, style of `\wrd{n}' (red line) appearing in (a) is  mimicked by HWT, for a different word including same character `\wrd{n}'. Similarly, a group of characters in `\wrd{thought}' and `\wrd{personalities}' (blue and magenta lines) are styled in a way that matches with words (`\wrd{throughout}' and `\wrd{qualities}') sharing some common characters in (a). Furthermore, HWT preserves cursive patterns and connectivity of all characters in word `\wrd{also}' (green line).}

 \vspace{-0.5cm}
  \label{fig:intro}
\end{figure}

Generating realistic synthetic handwritten text images, from typed text, that is versatile in terms of both writing style and lexicon is a challenging problem. Automatic handwritten text generation can be beneficial for people having disabilities or injuries that prevent them from writing, translating a note or a memo from one language to another by adapting an author's writing style or gathering additional data for training deep learning-based handwritten text recognition models. Here, we investigate the problem of realistic handwritten text generation of unconstrained text sequences with arbitrary length and diverse calligraphic attributes representing writing styles of a writer.



Generative Adversarial Networks (GANs)~\cite{goodfellow2014generative} have been investigated for offline handwritten text image generation \cite{chang2018generating,alonso2019adversarial, kang2020ganwriting, fogel2020scrabblegan, davis2020text}. These  methods  strive to directly synthesize text images by using offline handwriting images during training, thereby extracting useful features, such as writing appearance (\eg, ink width, writing slant) and line thickness changes. 
Alonso~\etal~\cite{alonso2019adversarial} propose a generative architecture that is conditioned on input content strings, thereby not restricted to a particular pre-defined vocabulary. However, their approach is trained on isolated fixed-sized word images and struggles to produce high quality arbitrarily long text along with suffering from style collapse. Fogel~\etal~\cite{fogel2020scrabblegan} introduce a ScrabbleGAN approach, where the generated image width is made proportional to the input text length. ScrabbleGAN is shown to achieve impressive results with respect to the content. However, both~\cite{alonso2019adversarial, fogel2020scrabblegan} do not adapt to a specific author's writing style.



Recently, GAN-based approaches~\cite{davis2020text, kang2020ganwriting} have been introduced for the problem of styled handwritten text image generation. These methods take into account both content \textit{and} style, when generating offline handwritten text images. 
Davis~\etal~\cite{davis2020text} propose an approach based on StyleGAN \cite{karras2019style} and learn generated handwriting image width based on style and input text. The GANwriting framework~\cite{kang2020ganwriting} conditions handwritten text generation process to both textual content and  style features in a few-shot setup.

In this work, we distinguish two key issues that impede the quality of styled handwritten text image generation in the existing GAN-based methods~\cite{davis2020text, kang2020ganwriting}. First, both style and content are loosely connected as their representative features are processed separately and later concatenated. While such a scheme enables entanglement between style and content at the word/line-level, it does not explicitly enforce style-content entanglement at the character-level.
Second, although these approaches capture global writing style (\eg, ink width, slant), they do not explicitly encode local style patterns (\eg, character style, ligatures). As a result of these issues, they struggle to accurately imitate local calligraphic style patterns from reference style examples (see Fig.~\ref{fig:intro}). Here, we look into an alternative approach that addresses both these issues in a single generative architecture.

\subsection{Contributions}
We introduce a new styled handwritten text generation approach built upon transformers, termed Handwriting Transformers (HWT),  that comprises an encoder-decoder network. 
The encoder network utilizes a  multi-headed self-attention mechanism to generate a self-attentive style feature sequence of a writer. This feature sequence is then input to the decoder network that consists of multi-headed self- and encoder-decoder attention to generate character-specific style attributes, given a set of query word strings. Consequently, the resulting output is fed to a convolutional decoder to generate final styled handwritten text image. Moreover, we improve the style consistency of the generated text by constraining the decoder output through a loss term whose objective is to re-generate style feature sequence of a writer at the encoder.





 
Our HWT imitates the style of a writer for a given query content through self- and encoder-decoder attention that emphasizes relevant self-attentive style features with respect to each character in that query. This enables us to capture style-content entanglement at the character-level. Furthermore, the self-attentive style feature sequence generated by our encoder captures both the global (\eg, ink width, slant ) \textit{and} local styles (\eg, character style, ligatures) of a writer within the feature sequence. 


We validate our proposed HWT by conducting extensive qualitative, quantitative and human-based evaluations. In the human-based evaluation, our proposed HWT was preferred 81\% of the time over recent styled handwritten text generation methods~\cite{davis2020text, kang2020ganwriting}, achieving human plausibility in terms of the writing style mimicry.  Following GANwriting~\cite{kang2020ganwriting}, we evaluate our HWT on all the four settings on the IAM handwriting dataset. On the extreme setting of out-of-vocabulary and unseen styles (OOV-U), where both query words and writing styles are never seen during training, the proposed HWT outperforms GANwriting~\cite{kang2020ganwriting} with an absolute gain of 16.5 in terms of Fr\`echet Inception Distance (FID) thereby demonstrating our generalization capabilities. Further, our qualitative analysis suggest that HWT performs favorably against existing works, generating realistic styled handwritten text images (see Fig.~\ref{fig:intro}).

\section{Related Work}
Deep learning-based handwritten text generation approaches can be roughly divided into stroke-based online and image-based offline methods. Online handwritten text generation methods \cite{graves2013generating, aksan2018deepwriting} typically require temporal data acquired from stroke-by-stroke recording of real handwritten examples (vector form) using a digital stylus pen. On the other hand, recent generative offline handwritten text generation methods \cite{chang2018generating,alonso2019adversarial, kang2020ganwriting, fogel2020scrabblegan} aim to directly generate text by performing training on offline handwriting images.

Graves~\cite{graves2013generating} proposes an approach based on Recurrent Neural Network (RNN) with Long-Term Memory (LSTM) cells, which enables predicting future stroke points from previous pen positions and an input text. Aksan \etal \cite{chang2018generating} propose a method based on conditional Variational RNN (VRNN), where the input is split into two separate latent variables to represent content and style. However, their approach tends to average out particular styles across writers, thereby reducing details~\cite{kotani2020generating}. In a subsequent work~\cite{aksan2019stcn}, the VRNN module is substituted by Stochastic Temporal CNNs which is shown to provide more consistent generation of handwriting. Kotani \etal \cite{kotani2020generating} propose an online handwriting stroke representation approach to represent latent style information by encoding writer-, character- and writer-character-specific style changes within an RNN model. 

Other than sequential methods, several recent works have investigated offline handwritten text image generation using GANs.  Haines \etal \cite{haines2016my} introduce an approach to generate new text in a distinct style inferred from source images. Their model requires a certain degree of human intervention during character segmentation and is limited to generating characters that are in the source images. The work of \cite{chang2018generating}  utilize  CycleGAN~\cite{zhu2017unpaired} to synthesize images of isolated handwritten characters of Chinese language. Alonso \etal \cite{alonso2019adversarial} propose an approach, where handwritten text generation is conditioned by character sequences. However, their approach suffers from style collapse hindering the diversity of synthesized images. Fogel \etal \cite{fogel2020scrabblegan} propose an approach, called ScrabbleGAN, that synthesizes handwritten word using a fully convolutional architecture. Here, the characters generated have similar receptive field width. A conversion model is introduced by \cite{mayr2020spatio} that approximates online handwriting from offline samples followed by using style transfer technique to the online data. This approach relies on conversion model's performance.

Few recent GAN-based works~\cite{davis2020text,kang2020ganwriting} investigate the problem of offline styled handwritten text image generation. Davis \etal \cite{davis2020text} propose an approach, where handwritten text generation is conditioned on both text and style, capturing global handwriting style variations. Kang \etal \cite{kang2020ganwriting} propose a method, called GANwriting, that conditions text generation on extracting style features in a few-shot setup  and textual content of a predefined fixed length. \\
\noindent\textbf{Our Approach:} Similar to GANwriting~\cite{kang2020ganwriting}, we also investigate the problem of styled handwritten text generation in a few-shot setting, where a limited number of style examples are available for each writer. Different from GANwriting, our approach possesses the flexibility to generate styled text of arbitrary length. In addition, existing works \cite{davis2020text, kang2020ganwriting} only capture style-content entanglement at the word/line-level. In contrast, our transformer-based approach enables style-content entanglement both at the word and character-level. While \cite{davis2020text, kang2020ganwriting} focuses on capturing the writing style at the global level, the proposed method strives to imitate both global and local writing style. 

\section{Proposed Approach}
\textbf{Motivation:} To motivate our proposed HWT method, 
we first distinguish two desirable characteristics to be considered when designing an approach for styled handwritten text generation with varying length and any desired style in a few-shot setting, without using character-level annotation.\\ 
\textbf{\emph{Style-Content Entanglement:}} 
As discussed earlier, both style and content are loosely connected in recently introduced GAN-based works \cite{kang2020ganwriting, davis2020text} with separate processing of style and content features, which are later concatenated. Such a scheme does not explicitly encode style-content entanglement at the character-level. Moreover, there are separate components for style, content modeling followed by a generator for decoding stylized outputs. In addition to style-content entanglement at word/line level, an entanglement between style and content at the character-level is expected to aid in imitating the character-specific writing style along with generalizing to out-of-vocabulary content. Further, such a tight integration between style and content leads to a cohesive architecture design.\\
\textbf{\emph{Global and Local Style Imitation:}} While the previous requisite focuses on connecting style and content, the second desirable characteristic aims at modeling both the global as well as local style features for a given calligraphic style.
Recent generative methods for styled handwritten text generation~\cite{kang2020ganwriting,davis2020text} typically capture the writing style at the global level (\eg, ink width, slant). However, the local style patterns (\eg, character style, ligatures) are not explicitly taken into account while imitating the style of a given writer. 
We argue that both global \textit{and} local style patterns are desired to be imitated for accurate styled text image generation.  
\subsection{Approach Overview}
\textbf{Problem Formulation:} We aim to learn the complex handwriting style characteristics of a particular writer $i\in \mathcal{W}$, where $\mathcal{W}$ includes a total of $M$ writers. We are given a set of $P$ handwritten word images, $\bm{X}_{i}^s = \left \{\bm{x}_{ij}  \right \}_{j=1}^{P}$, as few-shot calligraphic style examples of each writer. The superscript `$s$' in $\bm{X}_{i}^s$ denotes use of the set as a source of handwriting style which is transferred to the target images $\tilde{\bm{X}}_{i}^t$ with new textual content but consistent style properties.  The textual content is represented as a set of input query word strings $\mathcal{A} = \left \{\bm{a}_{j}  \right \}_{j=1}^{Q}$, where each word string $\bm{a}_{j}$ comprises an arbitrary number of characters from permitted characters set $\mathcal{C}$. The set $\mathcal{C}$ includes alphabets, numerical digits and punctuation marks \etc.
Given  a query text string $\bm{a}_j \in \mathcal{A}$ from an unconstrained set of vocabulary and $\bm{X}_{i}^s$, our model strives to generate new images $\tilde{\bm{X}}_{i}^t$ with the same text $\bm{a}_j$ in the writing style of a desired writer $i$.


 \begin{figure*}[t!]
\begin{center}
   \includegraphics[width=0.86\textwidth]{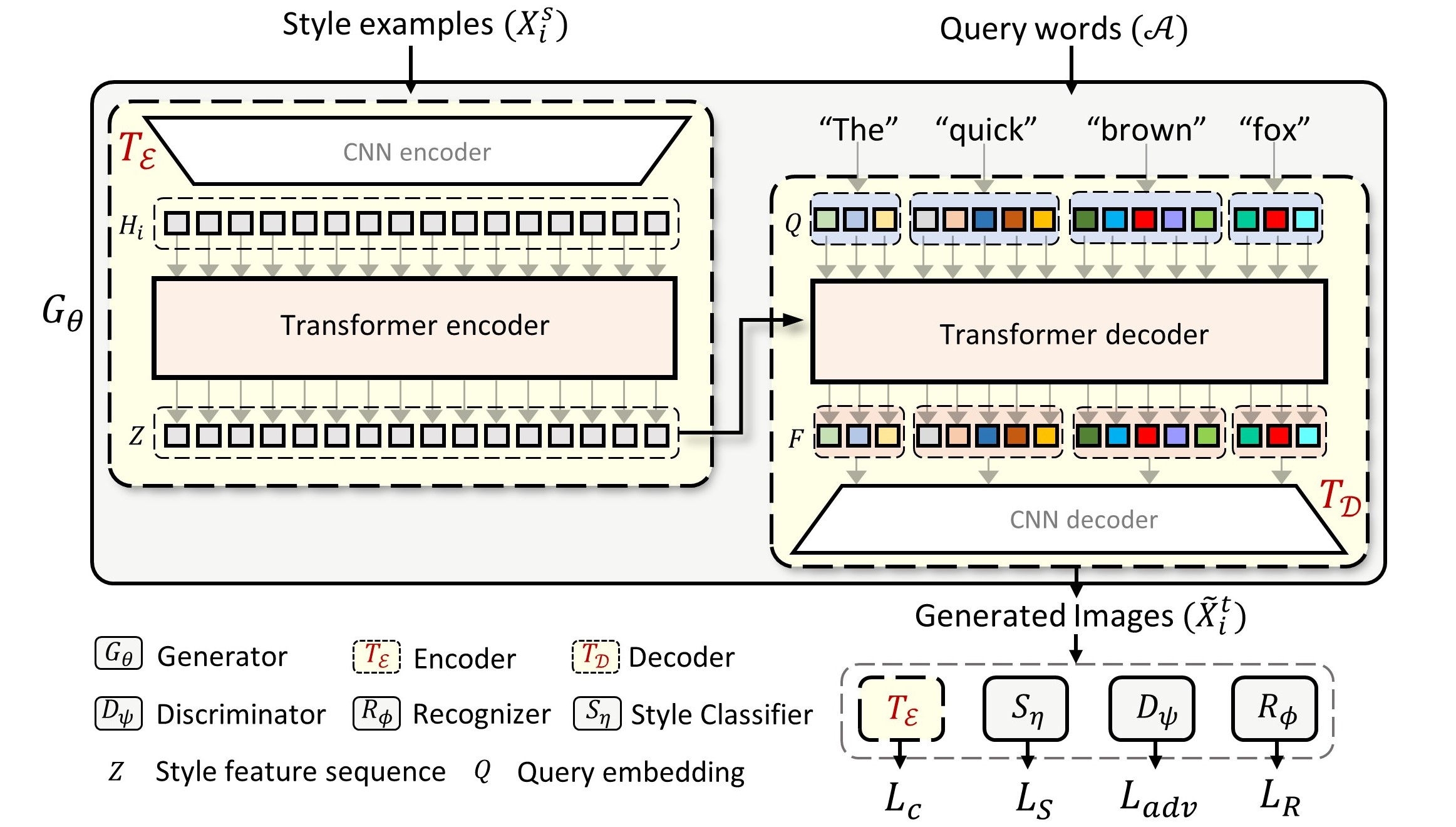}\vspace{-0.54cm}
\end{center}
\caption{Overall architecture of our Handwriting Transformers (HWT) to generate styled handwritten text images $\tilde{\bm{X}}_{i}^t$. HWT comprises a conditional generator having  an encoder $T_{\mathcal{E}}$ and a decoder network $T_{\mathcal{D}}$. Both the encoder and decoder networks constitute a hybrid convolution and multi-head self-attention design, which combines the strengths of CNN and transformer-based models \ie, highly expressive relationship modeling while working with limited handwriting style example images. Resultantly, our design seamlessly achieves style-content entanglement that encodes relationships between textual content and writer's style along with learning both global and local style patterns for given inputs ($\bm{X}_{i}^s$ and $\mathcal{A}$).} \vspace{-0.4cm}
\label{fig:method}
\end{figure*}


\textbf{Overall Architecture:} Fig.~\ref{fig:method} presents an overview of our proposed HWT approach, where a conditional generator $G_{\theta}$ synthesizes handwritten text images, a discriminator $D_{\psi}$ ensures realistic generation of handwriting styles, a recognizer $R_{\phi}$ aids in textual content preservation, and a style classifier $S_{\eta}$ ensures satisfactory transfer of the calligraphic styles. The focus of our design is the introduction of a transformer-based generative network for unconstrained styled handwritten text image generation. 
Our generator $G_{\theta}$ is designed in consideration to the desirable characteristics listed earlier leveraging the impressive learning capabilities of transformer models. To meticulously imitate a handwriting style, a model is desired to learn style-content entanglement as well as global and local style patterns. 

To this end, we introduce a transformer-based handwriting generation model, which enables us to capture the long and short range contextual relationships within the style examples $\bm{X}_{i}^s$ by utilizing a self-attention mechanism. In this way, both the global and local style patterns are encoded. Additionally, our transformer-based model comprises an encoder-decoder attention that allows style-content entanglement by inferring the style representation for each query character. A direct applicability of transformer-based design is infeasible in our few-shot setting due to its large data requirements and quadratic complexity. To circumvent this issue, our proposed architecture design utilizes the expressivity of a transformer within the CNN feature space.






The main idea of the proposed HWT method is simple but effective. A transformer-based encoder $T_{\mathcal{E}}$ is first used to model self-attentive style context that is later used by a decoder $T_{\mathcal{D}}$ to generate query text in a specific writer's style. We define learnable embedding vector $\bm{q}_{\wrd{c}} \in \mathbb{R}^{512}$ for each character $\wrd{c}$ of the permissible character set $\mathcal{C}$. For example, we represent the query word `\wrd{deep}' as a sequence of its respective character embeddings $\bm{Q}_{\wrd{deep}} = \left \{ \bm{q}_{\wrd{d}} \ldots \bm{q}_{\wrd{p}} \right \}$. We refer them as query embeddings. Such a character-wise representation of the query words and the transformer-based sequence processing helps our model to generate handwritten words of variable length, and also qualifies it to produce out-of-vocabulary words more efficiently. Moreover, it avoids averaging out individual character-specific styles in order to maintain the overall (global and local) writing style. The character-wise style interpolation and transfer is ensured by the self- and encoder-decoder attention in the transformer module that infers the style representation of each query character based on a set of handwritten samples provided as input. We describe the proposed generative architecture in Sec.~\ref{sec:generator} and the loss objectives in Sec.~\ref{sec:loss}.

\subsection{Generative Network}\label{sec:generator}
The generator $G_{\theta}$ includes two main components: an encoder network $T_{\mathcal{E}}:\bm{X}_{i}^s\rightarrow \bm{Z}$  and a decoder network $T_{\mathcal{D}}:(\bm{Z},\mathcal{A}) \rightarrow \tilde{\bm{X}}_{i}^t$.
The encoder produces a sequence of feature embeddings $\bm{Z} \in \mathbb{R}^{N\times d}$ (termed as style feature sequence)   
from a given set of style examples $\bm{X}_{i}^s$.  The decoder takes  $\bm{Z}$ as an input and  converts the input word strings $\bm{a}_{j} \in \mathcal{A}$ to realistic handwritten images $\tilde{\bm{X}}_{i}^t$ with same style as the given examples $\bm{X}_{i}^s$ of a writer $i$. 
Both the encoder and decoder networks constitute a \emph{hybrid} design based on convolution and multi-head self-attention networks. This design choice combines the strengths of CNNs and transformer models~\ie, highly expressive relationship modeling while working with limited handwriting images. Its worth mentioning that a CNN-only design would struggle to model long-term relations within sequences while an architecture based solely on transformer networks would demand large amount of data and longer training times~\cite{khan2021transformers}.

\textbf{Encoder $T_{\mathcal{E}}$.} The encoder aims at modelling  both global and local calligraphic style attributes (\ie, slant, skew, character shapes, ligatures, ink widths \etc) from the style examples $\bm{X}_{i}^s$. Before feeding style images to the highly expressive transformer architecture, we need to represent the style examples as a sequence. A straightforward way would be to flatten the image pixels into a 1D vector \cite{dosovitskiy2020image}. However, given the quadratic complexity of transformer models and their large data requirements, we find this to be infeasible. Instead, we use a CNN backbone network to obtain sequences of convolutional features from the style images. First, we use a ResNet18 \cite{he2016deep} model to generate lower-resolution activation maps $\bm{h}_{ij} \in \mathbb{R}^{h \times w \times d}$ for each style image $\bm{x}_{ij}$. Then, we flatten the spatial dimension of $\bm{h}_{ij}$ to obtain a sequence of feature maps of size $n \times d$, where $n=h\times w$. Each vector in the feature sequence represents a region in the original image and can be considered as the image descriptor for that particular region. After that, we concatenate the feature sequence vectors extracted from all style images together to obtain a single tensor $\bm{H}_{i} \in \mathbb{R}^{N\times d}$, where $N=n\times P$. 

The next step includes modeling the global and local compositions between all entities of the obtained  feature sequence $\bm{Z}$. A transformer-based encoder is employed for that purpose. The encoder has $L$ layers, where each layer has a standard architecture that consists of a multi-headed self-attention module and a Multi-layer Perceptron (MLP) block. At each layer $l$, the multi-headed self-attention maps the input sequence from the previous layer $\bm{H}^{l-1}$ into a triplet (key $\bm{K}$, query $\bm{Q}$, value $\bm{V}$) of intermediate representations given by,
\begin{align}
\bm{Q}=\bm{H}^{l-1}\bm{W}^{Q},  \bm{K}=\bm{H}^{l-1}\textbf{W}^{K}, \bm{V}=\bm{H}^{l-1}\textbf{W}^{V}, \notag
\end{align}
 where $\textbf{W}^{Q} \in \mathbb{R}^{N \times d_q}$, $\textbf{W}^{K} \in \mathbb{R}^{N \times d_k}$ and $\textbf{W}^{V} \in \mathbb{R}^{N \times d_v}$ are the learnable wight matrix for query, key and value respectively. For each head, the process is represented as,
\begin{equation}
    \bm{O}^{j} = \text{softmax}\left ( \frac{\bm{Q}\bm{K}^T}{\sqrt{d_k}} \right )\bm{V} \in  \mathbb{R}^{N \times d_v}, \quad j\in \{1,..,J\}.
\end{equation}
The concatenation of all $J$ head outputs $\bm{O}= [\bm{O}^{1}, \ldots, \bm{O}^{J}]$ is then fed through an MLP layer to obtain the output feature sequence $\bm{H}^{l}$ for the layer $l$. This update procedure is repeated for a total of $L$ layers, resulting in the final feature sequence $\bm{Z} \in \mathbb{R}^{N \times d}$. To retain information regarding the order of input sequences being supplied, we add fixed positional encodings ~\cite{vaswani2017attention} to the input of each attention layer.

\begin{figure}[t!]
  \includegraphics[width=\linewidth]{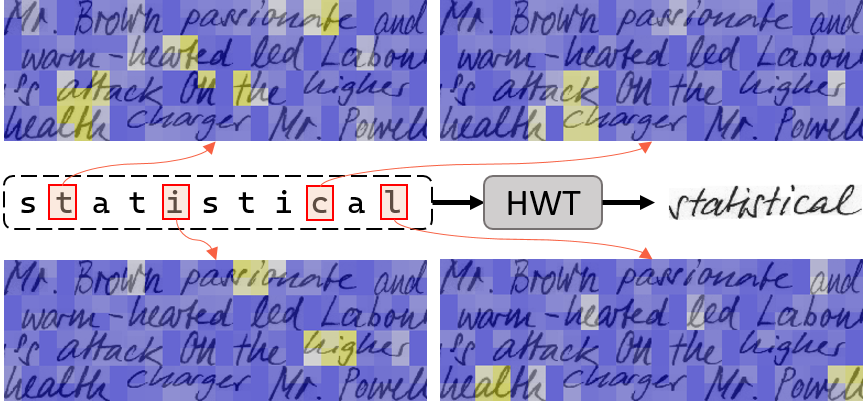} \vspace{-0.65cm}
  \caption{Visualization of encoder-decoder attention maps at the last layer of the transformer decoder. The attention maps are computed for each character in the query word (`\wrd{statistical}') which are then mapped to spatial regions (heat maps) in the example style images. Here, heat maps corresponding to the four different query characters `\wrd{t}', `\wrd{i}', `\wrd{c}' and `\wrd{l}' are shown. For instance, the top-left  attention map corresponding to the character `\wrd{t}', highlights multiple image regions containing the character `\wrd{t}'.} \vspace{-0.4cm}
  
  
  
  
  
  \label{fig:attention}
\end{figure}

\textbf{Decoder $T_{\mathcal{D}}$.} The initial stage in the decoder uses the standard architecture of the transformer that consists of multi-headed self- and encoder-decoder attention mechanisms. 
Unlike the self-attention, the encoder-decoder attention derives the key and value vectors from the output of the encoder, whereas the query vectors come from the decoder layer itself.
For an $m_j$ character word $\bm{a}_j \in \mathcal{A}$ (length $m_j$ being variable depending on the word), the query embedding $\bm{Q}_{\bm{a}_j} = \left \{ \bm{q}_{c_k} \right \}_{k=1}^{m_j}$ is used as a learnt positional encoding to each attention layer of the decoder. Intuitively, each query embedding learns to look up regions of interest in the style images to infer the style attributes of all query characters (see Fig.~\ref{fig:attention}). Over multiple consecutive decoding layers, these output embeddings accumulate style information, producing a final output $\bm{F}_{a_j} = \left \{ \bm{f}_{c_k}\right \}_{k=1}^{m_j} \in \mathbb{R}^{m_j \times d}$.  We  process the entire query embedding in parallel at each decoder layer. We add a randomly sampled noise vector $\mathcal{N}\left( 0,1 \right )$ to the output $\bm{F}_{\bm{a}_j}$ in order to model the natural variation of individual handwriting. For an $m$-character word, we concatenate these $m_j$ embedding vectors and pass them through a linear layer, resulting in an $m_j \times 8192$ matrix. After reshaping it to a dimension of $512\times 4\times 4m_j$, we
pass it through a CNN decoder having four residual blocks followed by a $\tanh$ activation layer to obtain final output images (styled hand written text images).


\subsection{Training and Loss Objectives}\label{sec:loss}
Our training algorithm follows the traditional GAN paradigm, where a discriminator network $D_{\psi}$ is employed to tell apart the samples generated from generator $G_{\theta}$ from the real ones. As the generated word images are of varying width, the proposed discriminator $D_{\psi}$ is also designed to be convolutional in nature. We use the hinge version of the adversarial loss \cite{lim2017geometric} defined as,
\begin{equation}
\begin{split}
L_{adv} = & \mathbb{E}\left [\max \left (1 - D_{\psi}(\bm{X}_{i}^s, 0)  \right )  \right ] + \\ 
& \mathbb{E}\left [\max\left ( 1 + D_{\psi}(G_{\theta}(\bm{X}_{i}^s, \mathcal{A})) , 0 \right )  \right ].
\end{split}
\end{equation}
While $D_{\psi}$ promotes real-looking images, it does not  preserve the content or the calligraphic styles. To preserve the textual content in the generated samples we use a handwritten recognizer network $R_{\phi}$ that  examines whether the generated samples are actually real text. The recognizer $R_{\phi}$ is inspired by CRNN \cite{shi2016end}. The CTC loss \cite{graves2006connectionist} is used to compare the recognizer output to the query words that were given as input to $G_{\theta}$. Recognizer $R_{\phi}$ is only optimized with real, labelled, handwritten samples, but it is used to encourage $G_{\theta}$ to produce readable text with accurate content. The loss is defined as, 
\begin{equation}
L_{R} = \mathbb{E}_{\bm{x}\sim \left \{ \bm{X}_{i}^s, \tilde{\bm{X}}_{i}^t \right \}}\left [ -\sum \log \left ( p\left ( y_{r}|R_{\phi}\left ( \bm{x} \right ) \right ) \right )   \right ].
\end{equation}
Here, $y_r$ is the  transcription string of $\bm{x}\sim \left \{ \bm{X}_{i}^s, \tilde{\bm{X}}_{i}^t \right \}$.

A style classifier network $S_{\eta}$ is employed to guide the network $G_{\theta}$ in producing samples conditioned to a particular writing style. The network $S_{\eta}$ attempts to predict the writer of a given handwritten image. The cross-entropy objective is applied as a loss function. $S_{\eta}$ is trained only on the real samples using the loss given below, 
\begin{equation}
L_{S} = \mathbb{E}_{x\sim \left \{ \bm{X}_{i}^s, \tilde{\bm{X}}_{i}^t \right \}}\left [ -\sum y_{i}  log \left ( S_{\eta} \left ( \bm{x} \right ) \right )   \right ].
\end{equation}

An important feature of our design is to utilize a cycle loss that ensures the encoded style features have cycle consistency. This loss function enforces the decoder to preserve the style information in the decoding process, such that the original style feature sequence can be reconstructed from the generated image. Given the generated word images $\tilde{\bm{X}}_{i}^t$, we use the encoder $T_{\mathcal{E}}$ to reconstruct the style feature sequence $\tilde{\bm{Z}}$. The cycle loss $L_c$ minimizes the error between the style feature sequence $\bm{Z}$ and its reconstruction $\tilde{\bm{Z}}$ by means of a $L_1$ distance metric,
\begin{equation}
L_{c} = \mathbb{E}\left [\left \|T_{\mathcal{E}}(\bm{X}_{i}^s) - T_{\mathcal{E}}(\tilde{\bm{X}}_{i}^t) \right \|_{1}  \right ].
\end{equation}
The cycle loss  imposes a regularization to the decoder  for consistently imitating the writing style in the generated styled text images. 
Overall, we train our HWT model in an end-to-end manner with the following loss objective,
\begin{equation}
    L_{total} = L_{adv} + L_{S}+ L_{R} +  L_{c}.
\end{equation}
We observe balancing the gradients of the network $S_{\eta}$ and $R_{\phi}$ is helpful in the training with our loss formulation.
Following \cite{alonso2019adversarial}, we normalize the $\nabla S_{\eta}$ and $\nabla R_{\phi}$ to have the same standard deviation ($\sigma$) as adversarial loss gradients, 
\begin{equation}
    \nabla S_{\eta} \leftarrow \alpha \left ( \frac{\sigma_{D}}{\sigma_{S}}. \nabla S_{\eta} \right ), 
    \nabla R_{\phi} \leftarrow \alpha \left ( \frac{\sigma_{D}}{\sigma_{R}}. \nabla R_{\phi} \right ).
\end{equation}
Here, $\alpha$ is a hyper-parameter that is fixed to $1$ during the training of our model. 

\section{Experiments}
We perform extensive experiments on IAM handwriting dataset~\cite{marti2002iam}. It consists of 9862 text lines with around 62,857 English words, written by 500 different writers. 
For thorough evaluation, we reserve an exclusive subset of 160 writers for testing, while images from the remaining 340 writers are used for our model training. 
In all our experiments, we  resize images to a fixed height of 64 pixels, while maintaining the aspect ratio of original image. 
For training, we use $P=15$ style example images, as in~\cite{kang2020ganwriting}. 
Both the transformer encoder and transformer decoder networks employ three attention layers ($L=3$) and each attention layer applies multi-headed attention having 8 attention heads ($J=8$). We set the embedding size $d$ to 512. In all experiments, we train our model for 4k epochs with a batch size of 8 on a single V100 GPU. Adam optimizer is employed during training with a learning rate of 0.0002.  

\begin{table} [t!]
\centering
\caption{\textbf{Comparison of the HWT with GANwriting \cite{kang2020ganwriting} and Davis \etal \cite{davis2020text}}   in terms of FID scores computed between the generated text images and real text images of the  IAM dataset. Our HWT performs favorably against \cite{kang2020ganwriting,davis2020text} in all four settings: In-Vocabulary words and seen style (IV-S), In-Vocabulary words and unseen style (IV-U), Out-of-vocabulary content and seen style (OOV-S) and Out-of-vocabulary content and unseen style (OOV-U). On the challenging setting of OOV-U, HWT achieves an absolute gain of $16.5$ in FID score, compared to GANwriting \cite{kang2020ganwriting}. Best results are in bold.
}\vspace{-0.2cm}
\resizebox{\linewidth}{!}{  
\begin{tabular}{|l|llll|} 
\hline 
  & IV-S $\downarrow$ &IV-U $\downarrow$ & OOV-S $\downarrow$ & OOV-U   $\downarrow$   \\ 
\hline \hline
GANwriting~\cite{kang2020ganwriting} & 120.07  & 124.30  & 125.87   & 130.68    \\ 
\hline
Davis~\etal~\cite{davis2020text} & 118.56  & 128.75  & 127.11   & 136.67            \\ 
\hline 
\textbf{HWT (Ours)} & \textbf{106.97}  & \textbf{108.84}  & \textbf{109.45}  & \textbf{114.10}   \\
\hline
\end{tabular} } \vspace{-0.4cm}
\label{tab:subset}
\end{table}




\begin{figure*}[t!]
  \includegraphics[width=\textwidth]{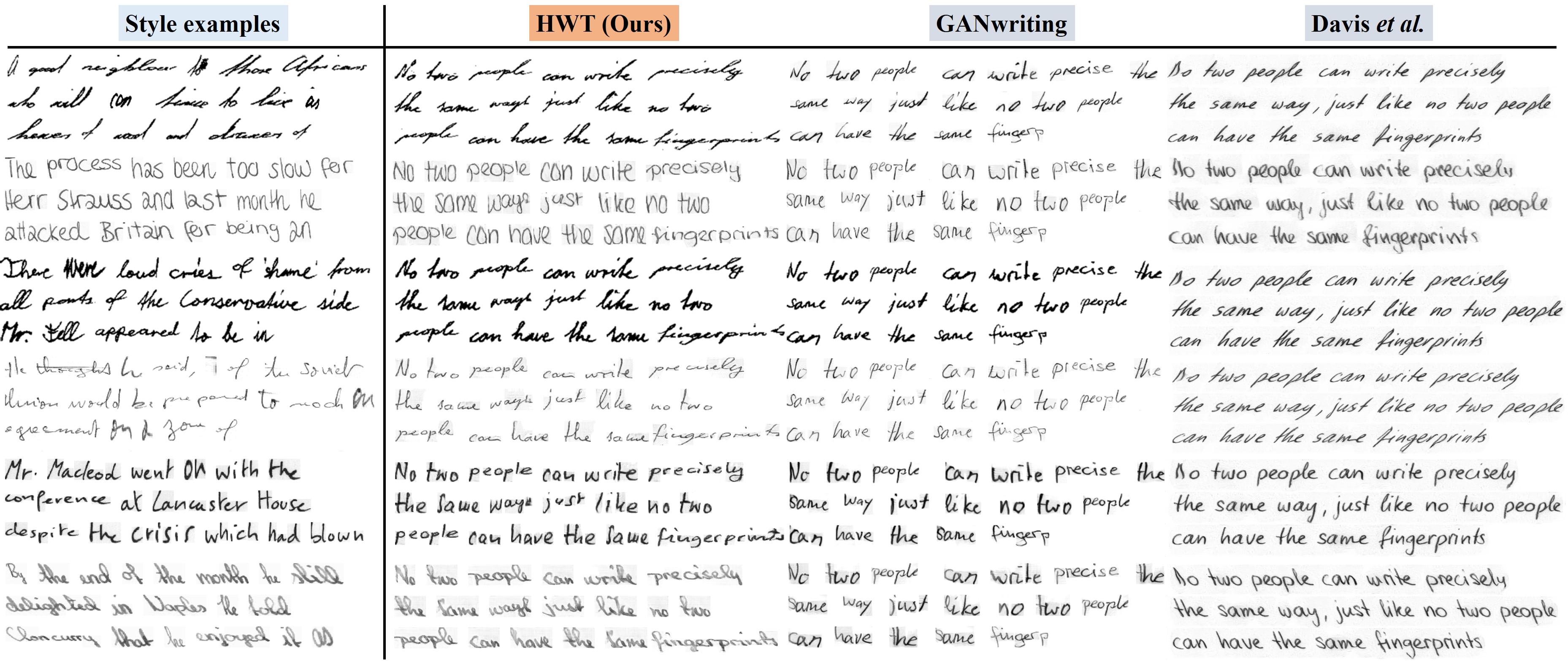}\vspace{-0.1cm}
  \caption{Qualitative comparison of our HWT (second column) with GANwriting \cite{kang2020ganwriting} (third column) and Davis~\etal~\cite{davis2020text} (fourth column).  We use the same textual content \emph{'No two people can write precisely the same way just like no two people can have the same fingerprints'} for all three methods. The first column shows the style examples from different writers. Davis~\etal~\cite{davis2020text} captures the global style, \eg slant, but struggles to mimic the character-specific style details. On the other hand, since GANwriting~\cite{kang2020ganwriting} is limited to a fixed length query words, it is unable to complete the provided textual content. Our HWT  better mimics global and local style patterns, generating more realistic handwritten text images. 
  }\vspace{-0.3cm}
  \label{fig:result}
\end{figure*}

\subsection{Styled Handwritten Text Generation}
We first evaluate (Tab.~\ref{tab:subset}) our approach for styled handwritten text image generation, where both style and content are desired to be imitated in the generated text image. 
Following~\cite{kang2020ganwriting}, we use Fr\`echet Inception Distance (FID)~\cite{heusel2017gans} evaluation metric for comparison. The FID metric is measured by computing the distance between the Inception-v3 features extracted from generated  and real samples for each writer and then averaging across all writers. We evaluate our HWT with GANwriting~\cite{kang2020ganwriting} and Davis \etal \cite{davis2020text} in four different settings: In-Vocabulary words and seen styles (IV-S), In-Vocabulary words and unseen styles (IV-U), Out-of-Vocabulary words and seen styles (OOV-S), and Out-of-Vocabulary words and unseen styles (OOV-U). Among these settings, most challenging one is the OOV-U, where both words and writing styles are never seen during training. For OOV-S and OOV-U settings, we use a set of 400 words that are distinct from IAM dataset transcription, as in~\cite{kang2020ganwriting}. In all four settings, the transcriptions of real samples and generated samples are different. Tab.~\ref{tab:subset} shows that HWT performs favorably against both existing methods~\cite{kang2020ganwriting, davis2020text}.   

Fig~\ref{fig:result} presents the qualitative comparison of HWT with ~\cite{kang2020ganwriting,davis2020text} for styled handwritten text generation. We present results for different writers, whose example style images are shown in the first column. For all the three methods, we use the same textual content.  
While Davis~\etal~\cite{davis2020text} follows the leftward slant of the last style example from the top, their approach struggles to capture character-level styles and cursive patterns (\eg see the word `\wrd{the}'). On the other hand, GANwriting~\cite{kang2020ganwriting} struggles to follow leftward slant of the last style example from the top and character-level styles. Our HWT better imitates both the global and local style patterns in these generated example text images.

\subsection{Handwritten Text Generation}
Here, we evaluate the quality of the handwritten text image generated by our HWT.  For a fair comparison with the recently introduced  ScrabbleGAN~\cite{fogel2020scrabblegan} and Davis~\etal~\cite{davis2020text}, we report our results in the \textit{same} evaluation settings as used by~\cite{fogel2020scrabblegan,davis2020text}. Tab.~\ref{tab:fullset} presents the comparison with  ~\cite{fogel2020scrabblegan,davis2020text} in terms of FID and geometric-score (GS).
Our HWT achieves favourable performance, compared to both approaches in terms of both FID and GS scores. Different from Tab.~\ref{tab:subset}, the results reported here in Tab.~\ref{tab:fullset} indicates the quality of the generated images, compared with the real examples in the IAM dataset, while ignoring style imitation capabilities.  


\begin{table}[t!]
\centering

\caption{\textbf{Handwritten text image generation quality comparison of our proposed HWT with ScrabbleGAN \cite{fogel2020scrabblegan} and Davis~\etal~\cite{davis2020text}} on the IAM dataset. Results are reported in terms of FID and GS by following the same evaluation settings, as in \cite{fogel2020scrabblegan,davis2020text}. Our HWT performs favorably against these methods in terms of both FID and GS. Best results are in bold.}\vspace{-0.2cm}
\resizebox{0.85\linewidth}{!}{  
\begin{tabular}{|l|l|l|} \hline
 & FID $\downarrow$ & GS $\downarrow$  \\                                            
\hline \hline
ScrabbleGAN \cite{fogel2020scrabblegan} & 20.72  & $2.56 \times 10^{-2}$    \\ 
\hline
\textcolor[rgb]{0.125,0.125,0.133}{Davis~\etal~\cite{davis2020text}}  & 20.65  & $4.88 \times 10^{-2}$   \\ 
\hline
\textbf{HWT (Ours)}  & \textbf{19.40}  & \boldsymbol{$1.01 \times 10^{-2}$}   \\
\hline
\end{tabular}}\vspace{-0.1cm}
\label{tab:fullset}
\end{table}

\begin{table}
\centering
\caption{\textbf{Impact of integrating transformer encoder (Enc), transformer decoder (Dec) and cycle loss (CL) to the baseline (Base)} on the OOV-U settings of IAM dataset. Results are reported in terms of FID score. Best results are reported in bold. On right, we show the effect of each component when generating two example words `\wrd{freedom}' and `\wrd{precise}' mimicking two given writing styles. 
} \vspace{0.0cm}
\resizebox{\linewidth}{!}{  
\begin{tabular}{|l|l|l|} 
\hline

\multirow{2}{*}{}                                                                               & \multirow{2}{*}{FID $\downarrow$}                 &    \multicolumn{1}{c|}{Style Example}  \\ \cline{3-3}  
 &   &   \raisebox{-.45\height}{\includegraphics[width=0.35\linewidth]{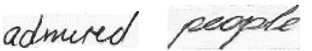}} \\  
\hline \hline
\raisebox{-.23\height}{Base}  &  \raisebox{-.23\height}{134.45} &    
\raisebox{-.45\height}{\includegraphics[width=0.35\linewidth]{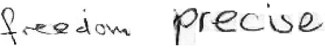}}
\\ 
\hline
\raisebox{-.23\height}{Base + Enc}        &    \raisebox{-.23\height}{128.80}                                       &       
\raisebox{-.45\height}{\includegraphics[width=0.35\linewidth]{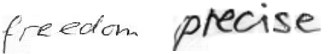}} \\ 
\hline
\raisebox{-.23\height}{Base + Dec}        &   \raisebox{-.23\height}{124.81}                                        &        
\raisebox{-.49\height}{\includegraphics[width=0.35\linewidth]{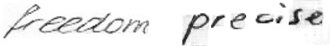}} \\  
\hline
\raisebox{-.33\height}{Base + Enc + Dec}  &     \raisebox{-.33\height}{116.50}                                     &    \raisebox{-.52\height}{\includegraphics[width=0.35\linewidth]{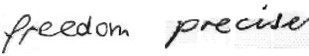}  }           \\ \hline
\raisebox{-.23\height}{Base + Enc + Dec + CL}&     \raisebox{-.23\height}{\textbf{114.10}}                                     &  \raisebox{-.53\height}{\includegraphics[width=0.35\linewidth]{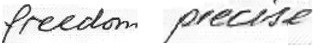}}               \\

\hline
\end{tabular}}\vspace{-0.4cm}
\label{tab:transformer1_ablation}
\end{table}

\subsection{Ablation study}
We perform multiple ablation studies on the IAM dataset to validate the impact of different components in our framework. 
Tab.~\ref{tab:transformer1_ablation} shows the impact of integrating transformer encoder (Enc), transformer decoder (Dec) and cycle loss (CL) to the baseline (Base).
Our baseline neither uses transformer modules nor utilizes cycle loss. It only employs a CNN encoder to obtain style features, whereas the content features are extracted from the one-hot representation of  query words. Both content and style features are passed through a CNN decoder to generate styled handwritten text images. While the baseline is able to generate realistic text images, it has a limited ability to mimic the given writer's style leading to inferior FID score (row 1). The introduction of the transformer encoder into the baseline (row 2) leads to an absolute gain of 5.6 in terms of FID score, highlighting the importance of our transformer-based self-attentive feature sequence in the generator encoder. We observe here that the generated sample still lacks details in terms of character-specific style patterns. When integrating the transformer decoder into the baseline (row 3), we observe a significant gain of 9.6 in terms of FID score.  
Notably, we observe a significant improvement (17.9 in FID) when integrating both transformer encoder and decoder to the baseline (row 4). This indicates the importance of 
self- and encoder-decoder attention for achieving realistic styled handwritten text image generation. The performance is further improved by the introduction of cycle loss to our final HWT architecture (row 4). 

As described earlier (Sec.~\ref{sec:generator}), HWT strives for style-content entanglement at character-level by feeding query character embeddings to the transformer decoder network. Here, we evaluate the effect of character-level content encoding (conditioning) by replacing it with word-level conditioning.
We obtain the word-level embeddings, by using an MLP that aims to obtain string representation of each query word. These embeddings are used as conditional input to the transformer decoder. Table~\ref{tab:cycle_ch_words} suggests that HWT benefits from character-level conditioning that ensures finer control of text style. The performance of word-level conditioning is limited to mimicking the global style, whereas our character-level approach ensures locally realistic as well as globally consistent style patterns.

\subsection{Human Evaluation}
Here, we present results of our two user studies on 100 human participants
to evaluate whether the proposed HWT achieves human plausibility in terms of the style mimicry. First, a \emph{User preference study} compares styled text images generated by our method with GANwriting~\cite{kang2020ganwriting} and Davis~\etal~\cite{davis2020text}. Second, 
a \emph{User plausibility study} that evaluates the proximity of the synthesized samples generated by our method to the real samples. In both studies, synthesized samples are generated using  \textit{unseen writing styles} of  test set writers  of IAM dataset, and  for textual content we use sentences from Stanford Sentiment Treebank \cite{socher2013recursive} dataset.



For \emph{User preference study}, each participant is shown the real handwritten paragraph of a person and synthesized handwriting samples of that person using HWT,  Davis~\etal~\cite{davis2020text} and  GANwriting~\cite{kang2020ganwriting}, randomly organized. The participants were asked to mark the best method for mimicking the real handwriting style. In total, we have collected 1000 responses. The results of this study shows that our proposed HWT was preferred 81\% of the time over the other two methods.



For \emph{User plausibility study}, each participant is shown a person's actual handwriting, followed by six samples, where each of these samples is either genuine or synthesized handwriting of the same person. Participants are asked to identify whether a given handwritten sample is genuine or not (forged/synthesized) by looking at the examples of the person's real handwriting. Thus, each participant provides 60 responses, thereby we collect 6000 responses for 100 participants. For this study, only 48.1\% of the images have been correctly classified, thereby showing a comparable performance to a random choice in a two-class problem.

\begin{table}[t!]
\centering
\caption{\textbf{Comparison between word and character-level conditioning} on IAM dataset. Results are reported in terms of FID score. Our character-level conditioning performs favorably, compared to its word-level counterpart. Best results are reported in bold. On the right, we show the effect of word and character-level conditioning, when generating two example words `\wrd{symbols}' and `\wrd{same}' mimicking  two given writing styles.}\vspace{0.0cm}
\resizebox{0.85\linewidth}{!}{
\begin{tabular}{|l|c|l|} 
\hline
\multirow{2}{*}{}                                                                             & \multirow{2}{*}{FID $\downarrow$}                 & \multicolumn{1}{c|}{Style Example}  \\ \cline{3-3}  
      &          &   \raisebox{-.52\height}{\includegraphics[width=0.37\linewidth]{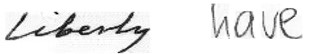}}             \\ 
\hline \hline
\multicolumn{1}{|c|}{\begin{tabular}[c]{@{}c@{}}\raisebox{-.48\height}{Word-level}  \end{tabular}}      & \multicolumn{1}{l|}{\raisebox{-.48\height}{126.87}}           &      \raisebox{-.58\height}{\includegraphics[width=0.37\linewidth]{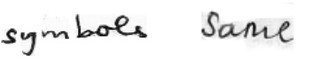}  }         \\ 
\hline
\multicolumn{1}{|c|}{\begin{tabular}[c]{@{}c@{}}\raisebox{-.44\height}{Character-level} \end{tabular}} & \multicolumn{1}{l|}{\textbf{\raisebox{-.44\height}{114.10}} } &    
\raisebox{-.47\height}{\includegraphics[width=0.37\linewidth]{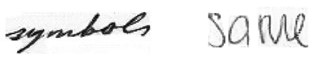}}  \\
\hline
\end{tabular}}\vspace{-0.4cm}
\label{tab:cycle_ch_words}
\end{table}

\section{Conclusion}
We introduced a transformer-based styled handwritten text image generation approach, HWT, that comprises a conditional generator having an encoder-decoder network. Our HWT captures the long and short range contextual relationships within the writing style example through a self-attention mechanism, thereby encoding both global and local writing style patterns. In addition, HWT utilizes an encoder-decoder attention that enables style-content entanglement at the character-level by inferring the style representation for each query character. Qualitative, quantitative and human-based evaluations show that our HWT produces realistic styled handwritten text images with varying length and any desired writing style. 

 


{\small
\bibliographystyle{ieee_fullname}
\bibliography{egbib}
}

\clearpage

\twocolumn[
  \begin{@twocolumnfalse}
    \begin{center}
        \section*{\Large{Handwriting Transformers\\Supplementary Material}}
        
        \vspace{1.5em}
    \end{center}
  \end{@twocolumnfalse}
]

\setcounter{equation}{0}
\setcounter{figure}{0}
\setcounter{table}{0}
\setcounter{page}{1}
\setcounter{section}{0}

In this supplementary material, we present additional human study details, additional qualitative results, and additional ablation study results. In Sec.~\ref{sec:human}, we provide  details of human study experiments.  Sec.~\ref{sec:attention} presents the additional visualisation results of transformer encoder-decoder attention maps.  Sec.~\ref{sec:qualitative} shows qualitative comparison of our proposed HWT. Sec.~\ref{sec:interpolation} shows the interpolations between two different calligraphic styles on the IAM dataset. Finally, Sec.~\ref{sec:ablation} presents additional ablation results.

\section{Human Study Additional Details}
\label{sec:human}  
\begin{figure*}[t!]
\begin{center}
\caption{Screenshot of the Interface used in \emph{User preference study} experiment. Each participant is shown the real handwritten text image (on the left side) of a person and synthesized handwriting text images (on the right side) of that person generated using three different methods. Participants have to mark the best method for mimicking the real handwriting style.}\vspace{-0.3cm}
   \includegraphics[width=1\textwidth]{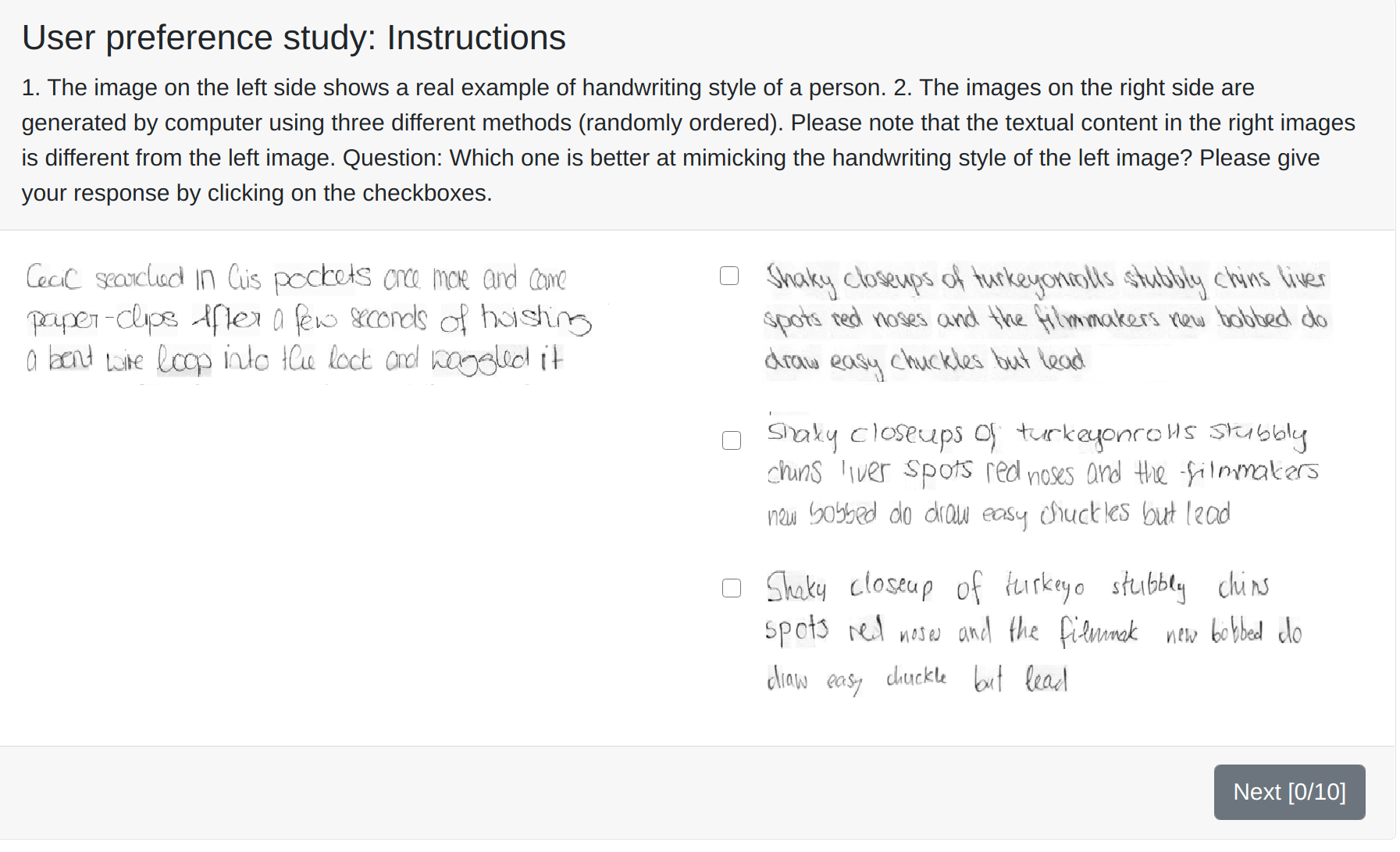}\vspace{-0.9cm}
  \label{fig:exp1}
\end{center}
\end{figure*}

Here, we present results of our two user studies on 100 human participants to evaluate the human plausibility in terms of style mimicry of our proposed HWT. In both these user studies, the forged samples are generated using  \textit{unseen writing styles} of test set writers  of IAM dataset, and  for textual content we use sentences from Stanford Sentiment Treebank \cite{socher2013recursive} dataset.\\
\noindent\textbf{User Preference Study:} Fig.~\ref{fig:exp1} shows the interface for the \emph{User preference study} experiment, which compares styled text images. In this study, each participant is shown a real handwritten text image of a person and the synthesized handwriting text images of that person using our proposed HWT,  Davis~\etal~\cite{davis2020text} and  GANwriting~\cite{kang2020ganwriting}. We randomly present generated results of these methods to the user. Then, the user can compare the real image and the generated images side by side on the same screen and without any time restriction to give the answer. Each participant is required to provide response for a total of ten questions. Overall, we have collected 1000 responses from 100 participants. Table~\ref{Tab:exp1} shows the results of \emph{User preference study}. Davis~\etal~\cite{davis2020text} and  GANwriting~\cite{kang2020ganwriting} were preferred 9\% (90 responses out of the total 1000) and 10\% (100 responses out of the total 1000), respectively.  Our proposed HWT was preferred 81\% (810 responses out of the total 1000 responses) over the other two existing methods.



\begin{figure*}[ht!]
\begin{center}
\caption{Screenshot of the Interface used in \emph{User plausibility study} experiment. Each participant is shown
a person’s actual handwriting (on the left side), followed by six samples (on the right side), where three out of these samples are genuine and the rest are synthesized. Participants have to classify each sample as genuine or forgery by looking at the real image. 
}\vspace{-0.3cm}
   \includegraphics[width=1\textwidth]{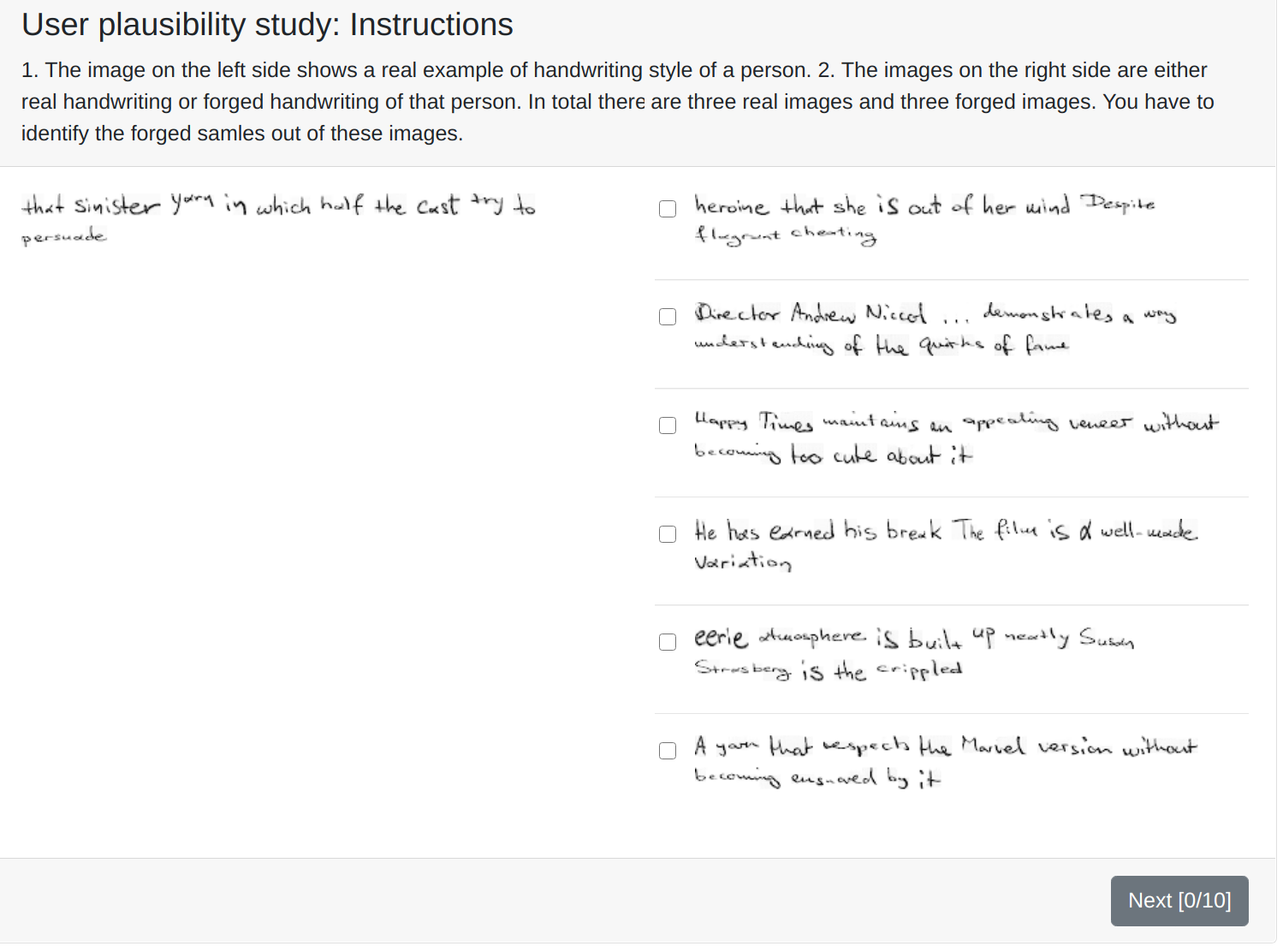}\vspace{-0.74cm}
   \label{fig:exp2}
\end{center}
 \end{figure*}

\noindent \textbf{User Plausibility Study:} Fig.~\ref{fig:exp2} shows the interface for the \emph{User plausibility study}, which evaluates the proximity of the synthesized samples generated by our proposed HWT to the real samples. Here, each participant is shown a person's actual handwriting, followed by six samples, where each of these samples is either genuine or synthesized handwriting of the same person. Participants are asked to identify whether a given handwritten sample is genuine or not (forged/synthesized) with no time limit restriction to answer the question. In total, we collect 6000 responses for 100 human participants as each one provides 60 responses. The study revealed that the generated images produced by our proposed HWT were deemed plausible. Table~\ref{Tab:exp2} shows the confusion matrix of the human assessments. For this particular study, only 48.1\% of the images have been correctly classified, which indicates a comparable performance to random choice in a two-class problem.


\begin{table}[t!]
\centering
\caption{\emph{User preference study} in comparison to GANwriting \cite{kang2020ganwriting} and Davis~\etal~\cite{davis2020text}. The result shows that our proposed HWT was preferred 81\% of the time over the other two compared methods. }
\label{tab:userexp1}
\begin{tabular}{|l|c|c|} 
\hline
           & \begin{tabular}[c]{@{}c@{}}Total \\Responses \end{tabular} & \begin{tabular}[c]{@{}c@{}}User \\Preferences \end{tabular}  \\ 
\hhline{|===|}
GANwriting~\cite{kang2020ganwriting} & \multirow{3}{*}{1000}                                      & 100                                                               \\ 
\cline{1-1}\cline{3-3}
Davis~\etal~\cite{davis2020text}      &                                                            & 90                                                                \\ 
\cline{1-1}\cline{3-3}
HWT (Ours) &                                                            & 810                                                                \\
\hline

\end{tabular} \vspace{-0.1cm}
\label{Tab:exp1}
\end{table}

\begin{table}[t!]
\centering\setlength{\tabcolsep}{10pt}
\caption{Confusion matrix (\%) obtained from \emph{User plausibility study}. Only 48.1\% of the images were correctly classified, indicating an output comparable to a random choice in a two-class problem.}
\label{tab:userexp2}
\begin{tabular}{|l|ll|} 
\hline
\multirow{2}{*}{Actual} & \multicolumn{2}{c|}{Predicted}   \\ 
\cline{2-3}
                        & Real                     & Fake  \\ 
\hhline{|===|}
Real                    & \multicolumn{1}{c}{24.9} & 25.1  \\ 
\hline
Fake                    & 26.8                     & 23.2  \\
\hline
\end{tabular}\vspace{-0.3cm}
\label{Tab:exp2}
\end{table}

\section{Additional Visualizations of Transformer Encoder-Decoder Attention}
\label{sec:attention}

Fig.~\ref{fig:attention} shows the visualization of attention maps obtained using  encoder-decoder of our approach (HWT) at the last layer of the transformer decoder. We compute the attention matrices for four different words: `\wrd{laughs}', `\wrd{because}', `\wrd{inside}', and `\wrd{fashion}'. Note that the attention maps generated by our model focus on the relevant regions of interest in the style examples for each query character. For instance, to infer character-specific style attributes of a given character `\wrd{h}' in the query word `\wrd{laughs}', the model gives priority to multiple image regions containing the character `\wrd{h}'. Note that if the query character isn't found in the style examples, the model attempts to find similar characters. For example, to obtain character representation of `\wrd{u}' in the query word `\wrd{laughs}', the attention algorithm highlights image regions containing similar characters (\eg `\wrd{n}').



\section{Additional Qualitative Comparison}
\label{sec:qualitative}
\begin{figure*}[t!]
\begin{center}
\caption{Additional qualitative comparisons of our proposed HWT with GANwriting \cite{kang2020ganwriting} and Davis~\etal~\cite{davis2020text}, when generating the same text \emph{`With more character development this might have been an eerie thriller with better payoffs it could have been a thinking'}. }
   \includegraphics[width=.95\textwidth]{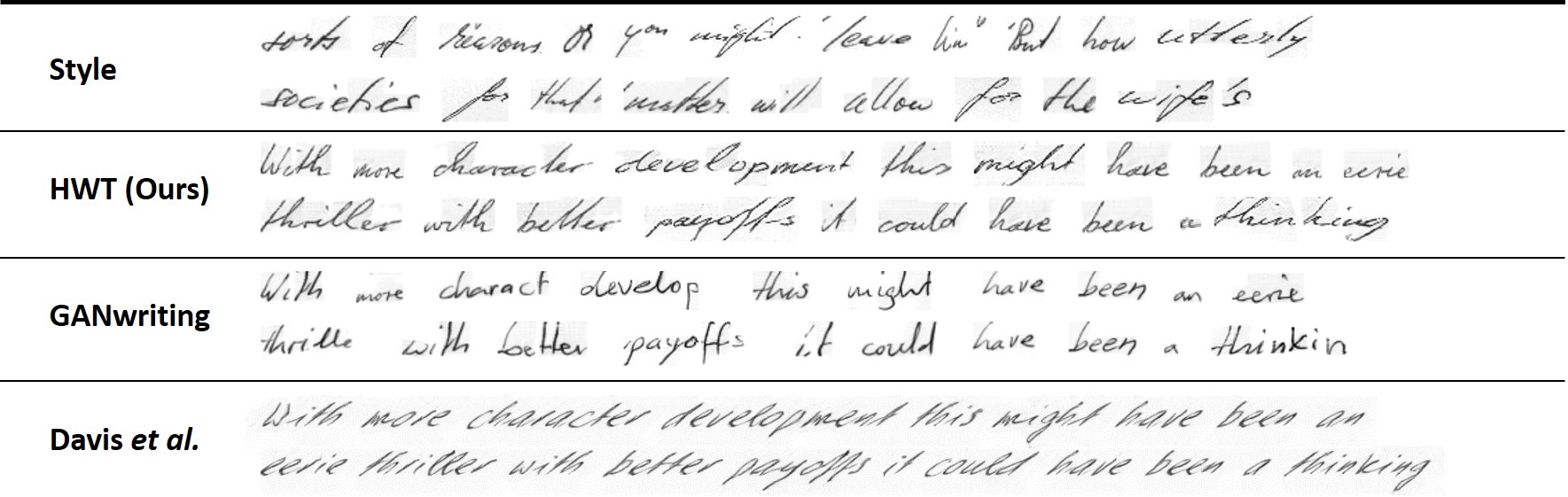}\vspace{-0.54cm}
   \label{fig:res1}
\end{center}
 \end{figure*}

\begin{figure*}[t!]
\begin{center}
\caption{Additional qualitative comparisons of our proposed HWT with GANwriting~\cite{kang2020ganwriting} and Davis~\etal~\cite{davis2020text}, when generating the same text \emph{`Its not helpful to listen to extremist namecalling regardless of whether you think Kissinger was a calculating'}.}
   \includegraphics[width=.95\textwidth]{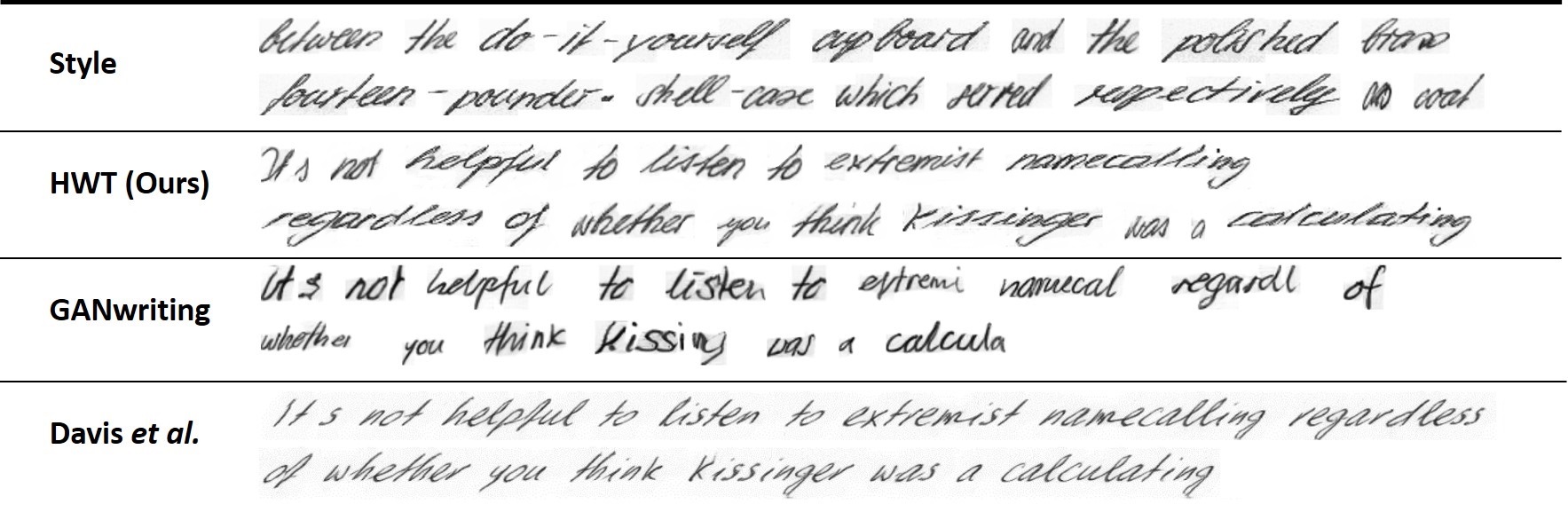}\vspace{-0.54cm}
    \label{fig:res2}
\end{center}
 \end{figure*}

\begin{figure*}[t!]
\begin{center}
\caption{Additional qualitative comparisons of our proposed HWT with GANwriting \cite{kang2020ganwriting} and Davis~\etal~\cite{davis2020text}, when generating the same text \emph{`Shaky closeups of turkeyonrolls stubbly chins liver spots red noses and the filmmakers new bobbed do draw easy chuckles but'}. }
   \includegraphics[width=.95\textwidth]{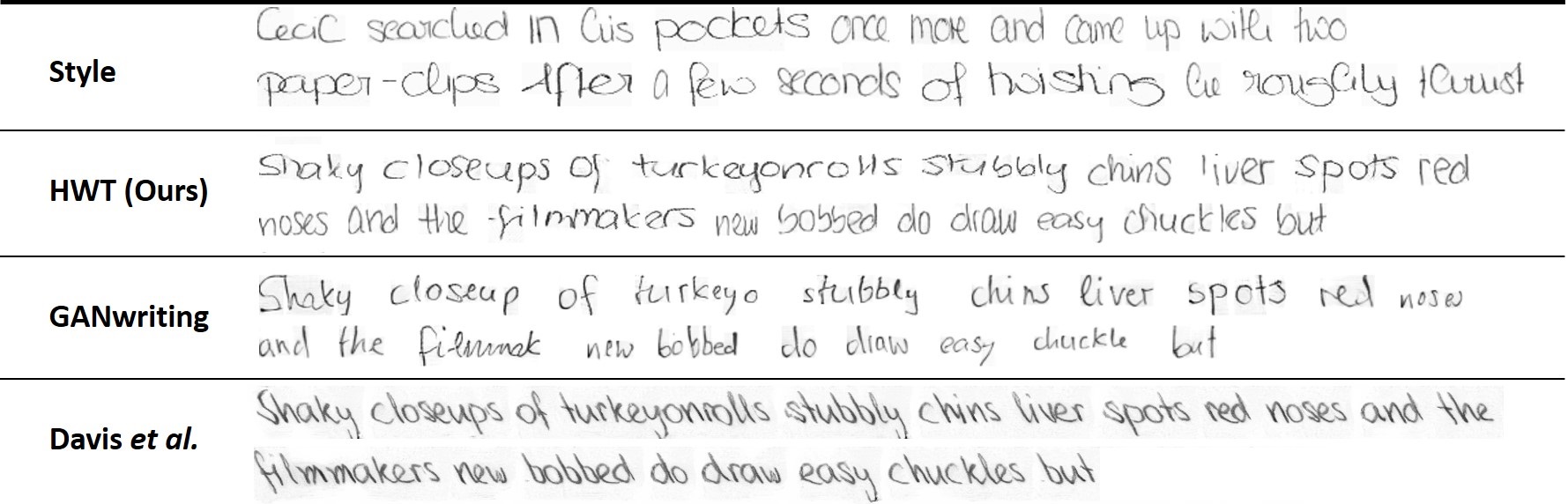}\vspace{-0.54cm}
    \label{fig:res3}
\end{center}
 \end{figure*}

\begin{figure*}[t!]
\begin{center}
\caption{Additional qualitative comparisons of our proposed HWT with GANwriting \cite{kang2020ganwriting} and Davis~\etal~\cite{davis2020text}, when generating the same text \emph{`This film was made by and for those folks who collect the serial killer cards and are fascinated by the mere suggestion'}. }
   \includegraphics[width=.95\textwidth]{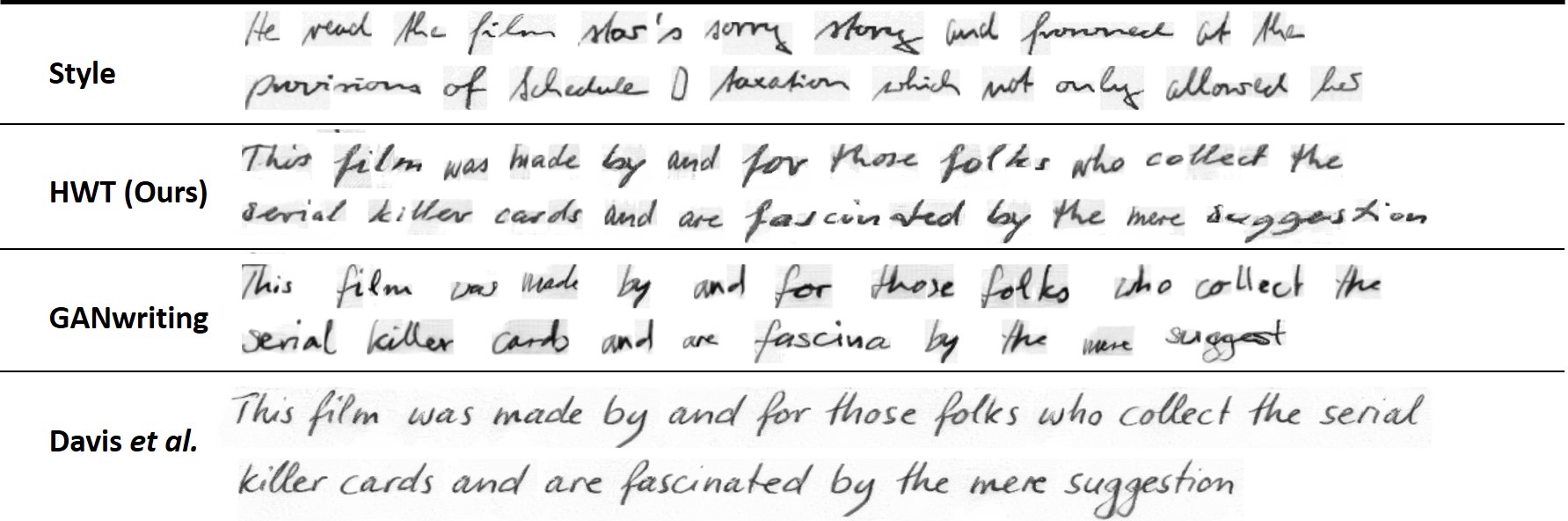}\vspace{-0.54cm}
    \label{fig:res4}
\end{center}
 \end{figure*}

\begin{figure*}[t!]
\begin{center}
\caption{Additional qualitative comparisons of our proposed HWT with GANwriting \cite{kang2020ganwriting} and Davis~\etal~\cite{davis2020text}, when generating the same text \emph{`Its a drawling slobbering lovable runon sentence of a film a Southern Gothic with the emotional arc of its raw blues'}. }
   \includegraphics[width=.95\textwidth]{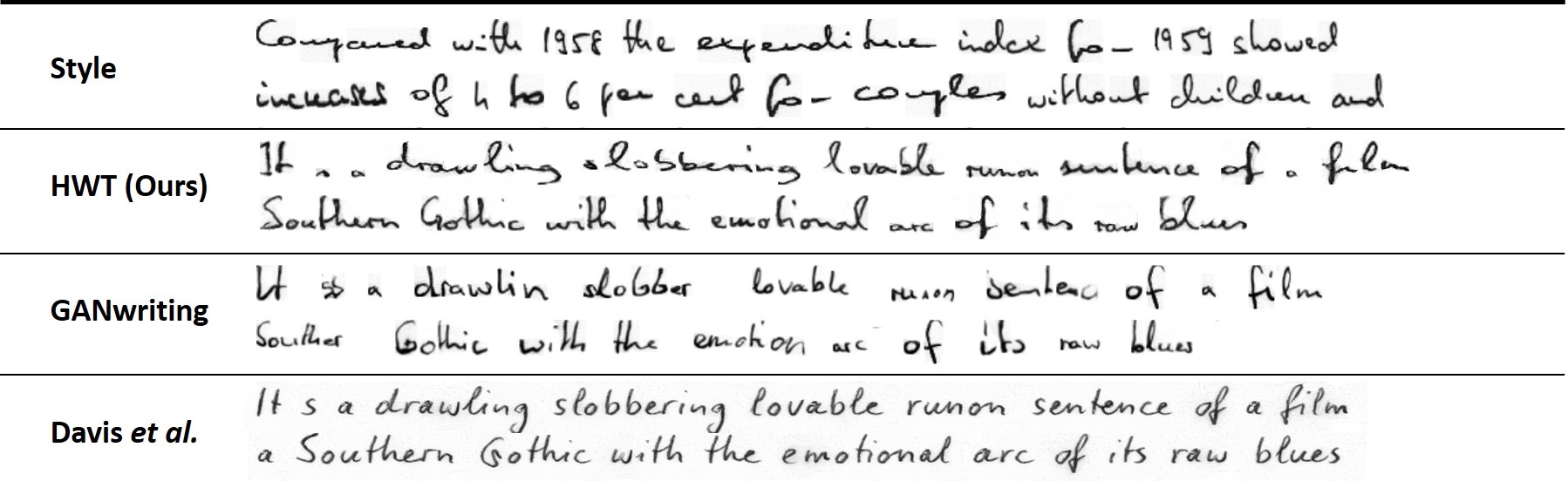}\vspace{-0.54cm}
    \label{fig:res5}
\end{center}
 \end{figure*}

\begin{figure*}[t!]
\begin{center}
\caption{Additional qualitative comparisons of our proposed HWT with GANwriting \cite{kang2020ganwriting} and Davis~\etal~\cite{davis2020text}, when generating the same text \emph{`LRB W RRB hile long on amiable monkeys and worthy environmentalism Jane Goodalls Wild Chimpanzees is short'}. }
   \includegraphics[width=.95\textwidth]{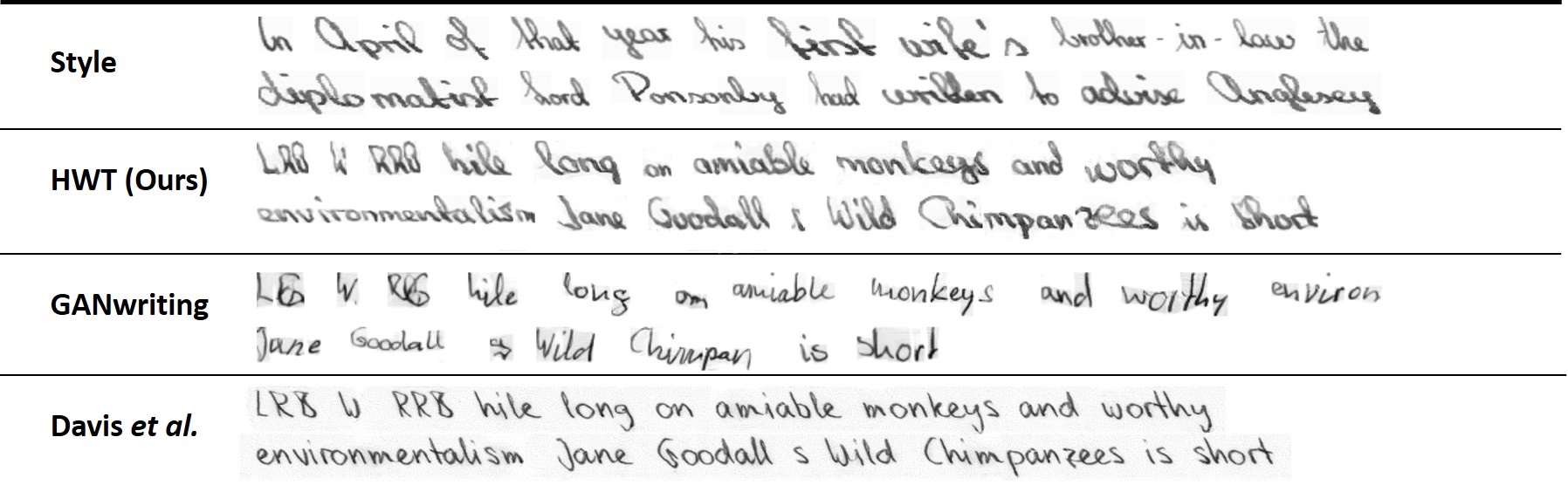}\vspace{-0.54cm}
    \label{fig:res6}
\end{center}
 \end{figure*}

\begin{figure*}[t!]
\begin{center}
\caption{Additional qualitative comparisons of our proposed HWT with GANwriting \cite{kang2020ganwriting} and Davis~\etal~\cite{davis2020text}, when generating the same text \emph{`For close to two hours the audience is forced to endure three terminally depressed mostly inarticulate hyper dysfunctional'} }
   \includegraphics[width=.95\textwidth]{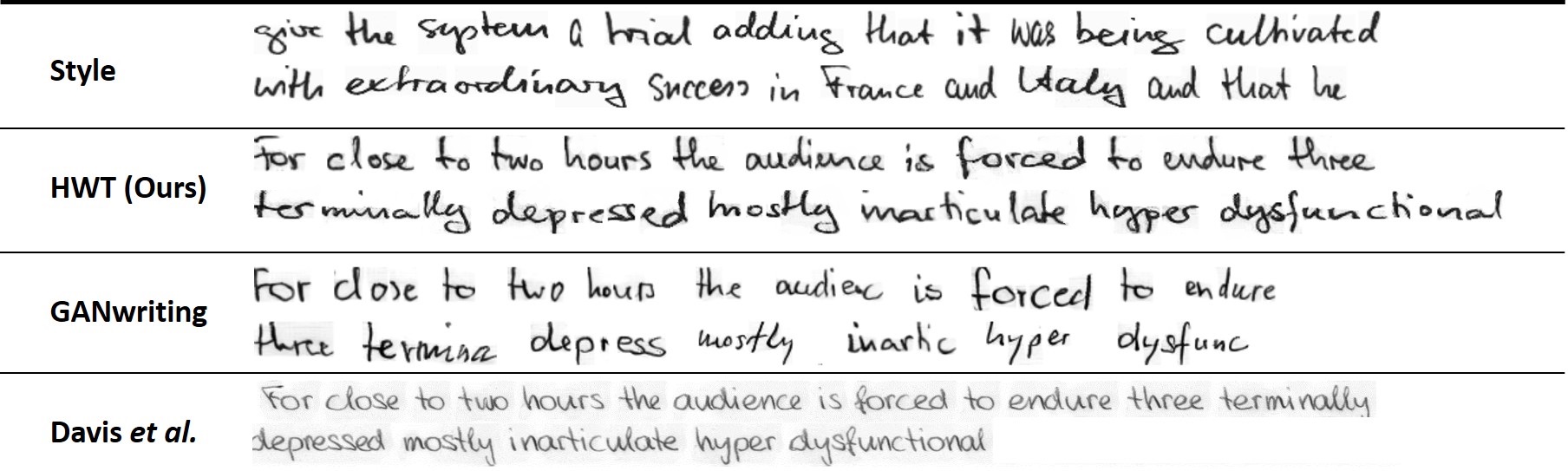}\vspace{-0.54cm}
    \label{fig:res7}
\end{center}
 \end{figure*}

\begin{figure*}[t!]
\begin{center}
\caption{Additional qualitative comparisons of our proposed HWT with GANwriting \cite{kang2020ganwriting} and Davis~\etal~\cite{davis2020text}, when generating the same text \emph{`Claude Chabrols camera has a way of gently swaying back and forth as it cradles its characters veiling tension beneath'}. }
   \includegraphics[width=.95\textwidth]{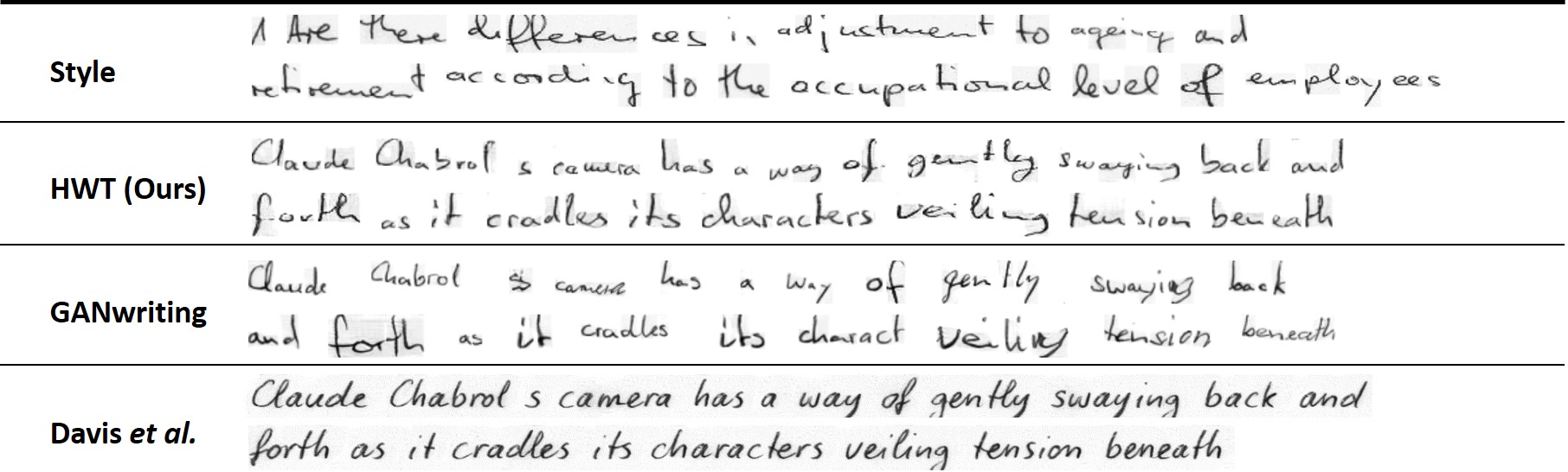}\vspace{-0.54cm}
    \label{fig:res8}
\end{center}
 \end{figure*}

\begin{figure*}[t!]
\begin{center}
\caption{Additional qualitative comparisons of our proposed HWT with GANwriting \cite{kang2020ganwriting} and Davis~\etal~\cite{davis2020text}, when generating the same text \emph{`Though the plot is predictable the movie never feels formulaic because the attention is on the nuances of the'}. }
   \includegraphics[width=.95\textwidth]{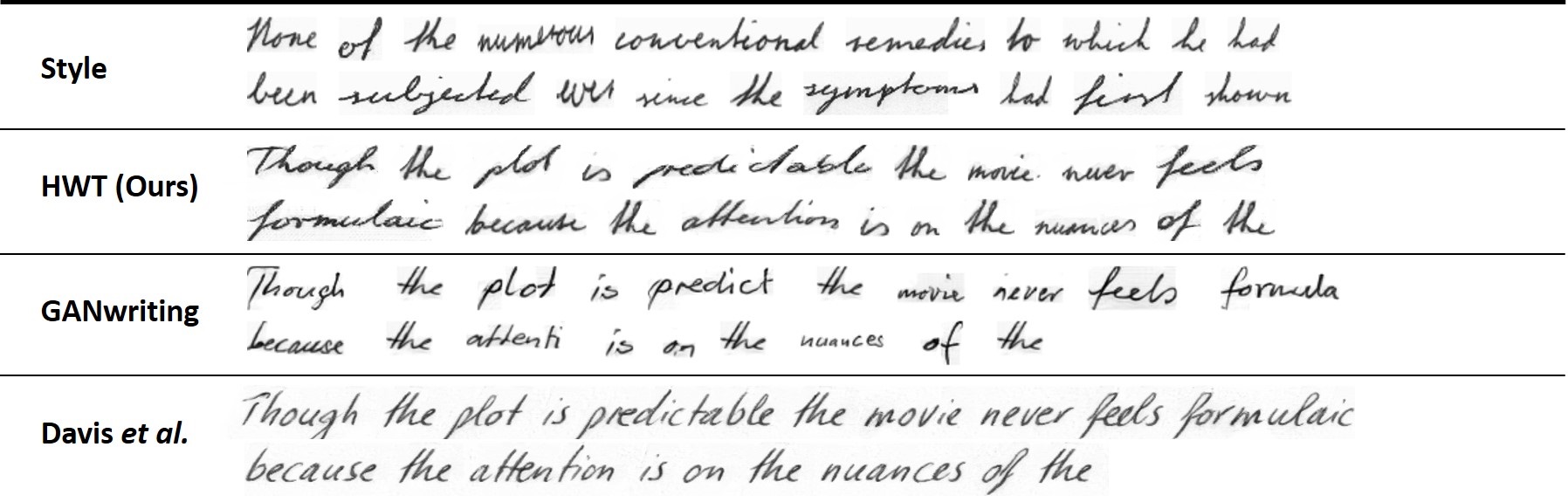}\vspace{-0.54cm}
    \label{fig:res9}
\end{center}
 \end{figure*}

\begin{figure*}[t!]
\begin{center}
\caption{Additional qualitative comparisons of our proposed HWT with GANwriting \cite{kang2020ganwriting} and Davis~\etal~\cite{davis2020text}, when generating the same text \emph{`A comingofage tale from New Zealand whose boozy languid air is balanced by a rich visual clarity and deeply felt'}. }
   \includegraphics[width=.95\textwidth]{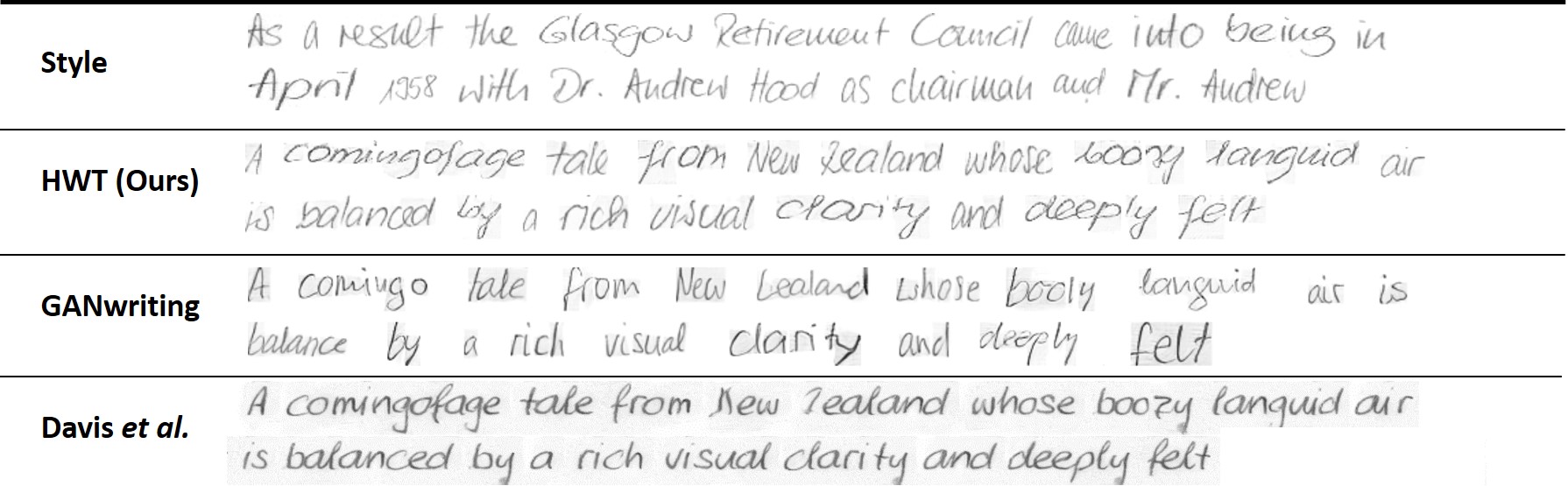}\vspace{-0.54cm}
    \label{fig:res10}
\end{center}
 \end{figure*}

\begin{figure*}[t!]
\begin{center}
\caption{Additional qualitative comparisons of our proposed HWT with GANwriting \cite{kang2020ganwriting} and Davis~\etal~\cite{davis2020text}, when generating the same text \emph{`Unfortunately Kapur modernizes AEW. Masons story to suit the sensibilities of a young American a decision that plucks The'}. }
   \includegraphics[width=.95\textwidth]{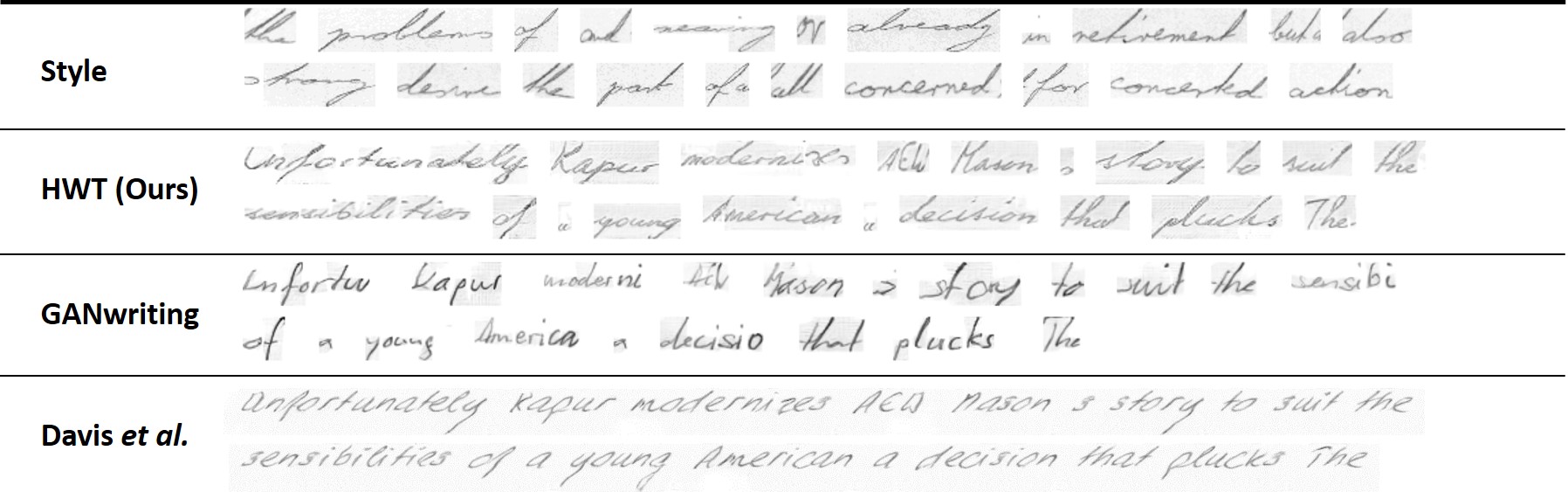}\vspace{-0.54cm}
    \label{fig:res11}
\end{center}
 \end{figure*}

\begin{figure*}[t!]
\begin{center}
\caption{Additional qualitative comparisons of our proposed HWT with GANwriting \cite{kang2020ganwriting} and Davis~\etal~\cite{davis2020text}, when generating the same text \emph{`Unless Bob Crane is someone of particular interest to you this films impressive performances and adept direction are'}.}
   \includegraphics[width=.95\textwidth]{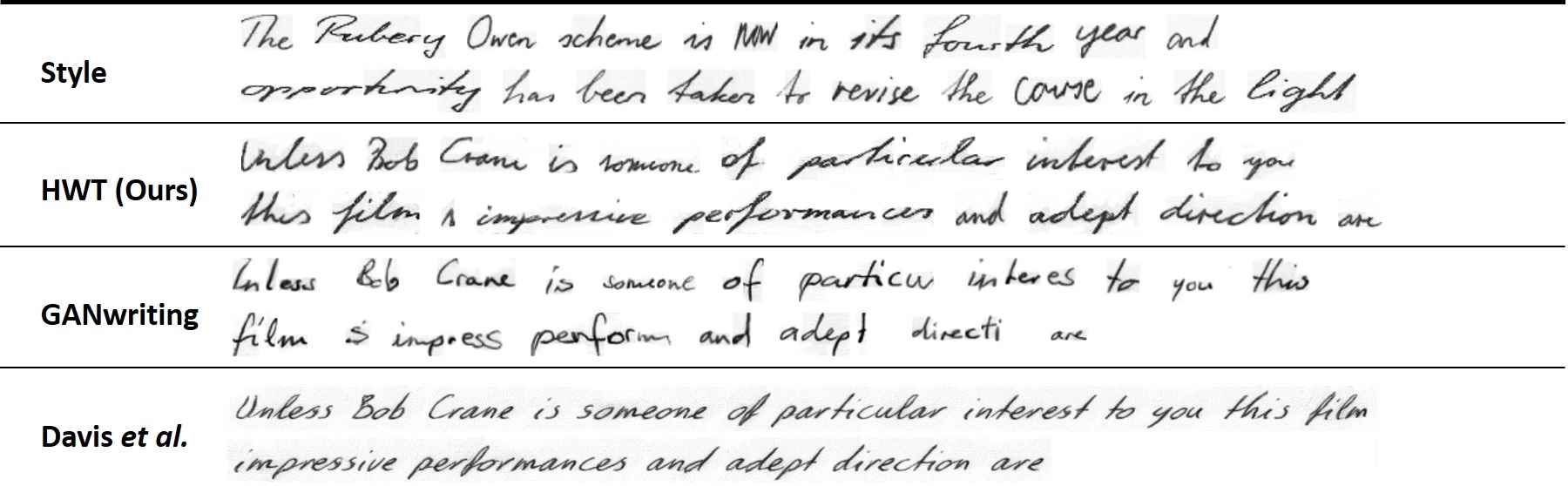}\vspace{-0.54cm}
    \label{fig:res12}
\end{center}
 \end{figure*}

\begin{figure*}[t!]
\begin{center}
\caption{Additional qualitative comparisons of our proposed HWT with GANwriting \cite{kang2020ganwriting} and Davis~\etal~\cite{davis2020text}, when generating the same text \emph{`Affirms the gifts of all involved starting with Spielberg and going right through the ranks of the players oncamera and off'}. }
   \includegraphics[width=.95\textwidth]{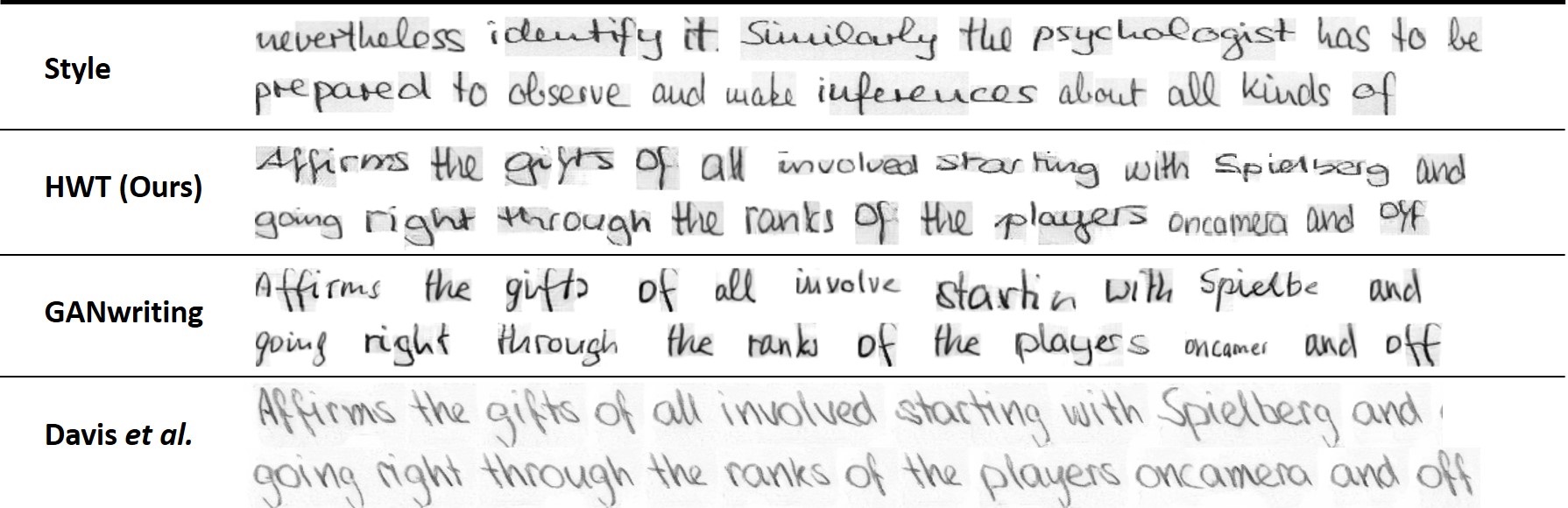}\vspace{-0.54cm}
    \label{fig:res13}
\end{center}
 \end{figure*}

\begin{figure*}[t!]
\begin{center}
\caption{Additional qualitative comparisons of our proposed HWT with GANwriting \cite{kang2020ganwriting} and Davis~\etal~\cite{davis2020text}, when generating the same text \emph{`Though this rude and crude film does deliver a few gut-busting laughs its digs at modern society are all things we ve seen'}. }
   \includegraphics[width=.95\textwidth]{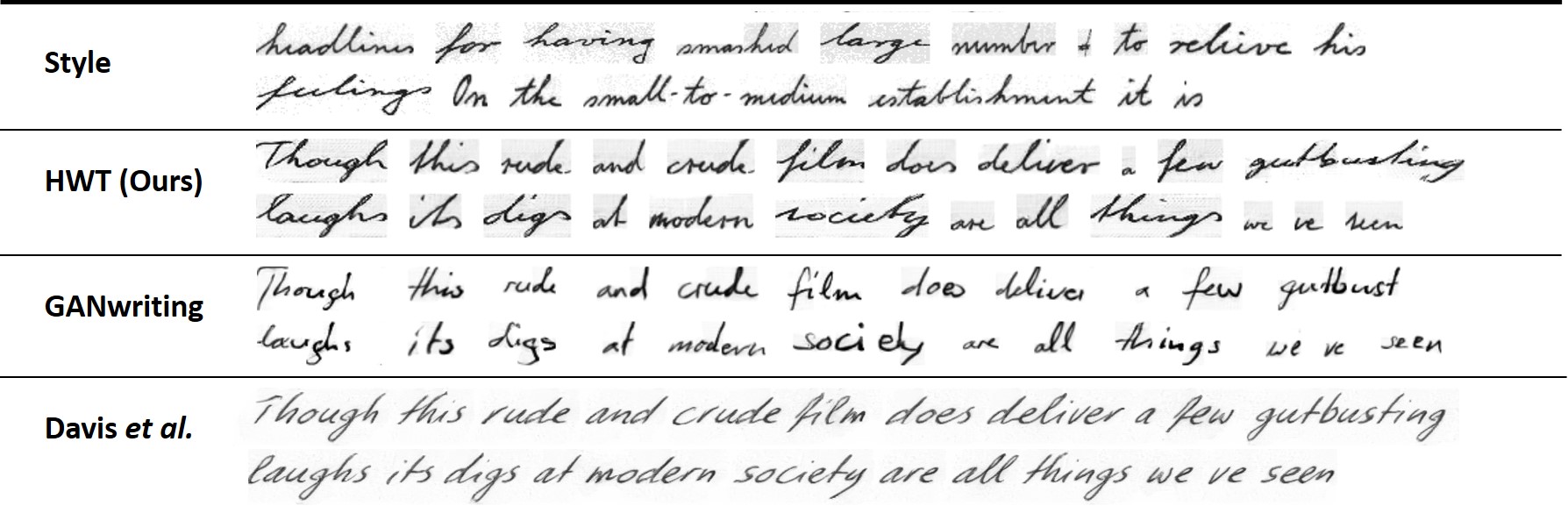}\vspace{-0.54cm}
   \label{fig:res14}
\end {center}
 \end{figure*}

\begin{figure*}[t!]
\begin{center}
\caption{Additional qualitative comparisons of our proposed HWT with GANwriting \cite{kang2020ganwriting} and Davis~\etal~\cite{davis2020text}, when generating the same text \emph{`You ll laugh at either the obviousness of it all or its stupidity or maybe even its inventiveness but the point is'}. }
   \includegraphics[width=.95\textwidth]{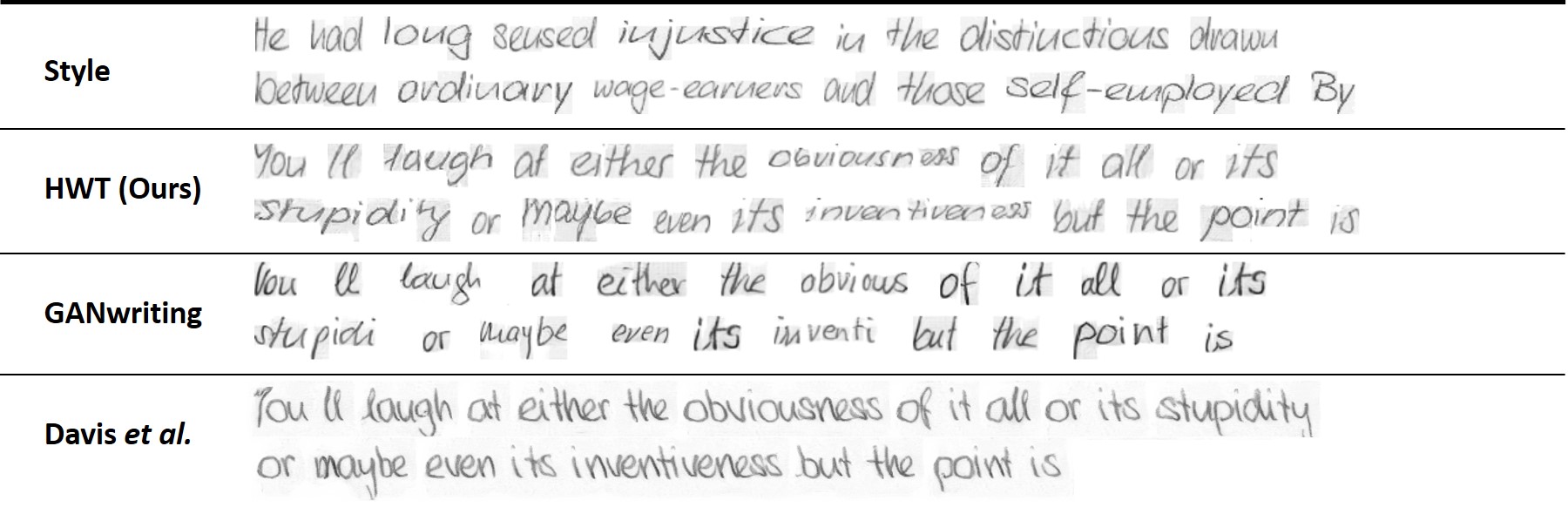}\vspace{-0.54cm}
   \label{fig:res15}
\end{center}
 \end{figure*}

\begin{figure*}[t!]
\begin{center}
\caption{Additional qualitative comparisons of our proposed HWT with GANwriting \cite{kang2020ganwriting} and Davis~\etal~\cite{davis2020text}, when generating the same text \emph{`Writerdirector s Mehta s effort has tons of charm and the whimsy is in the mixture the intoxicating masala of cultures'}. }
   \includegraphics[width=.95\textwidth]{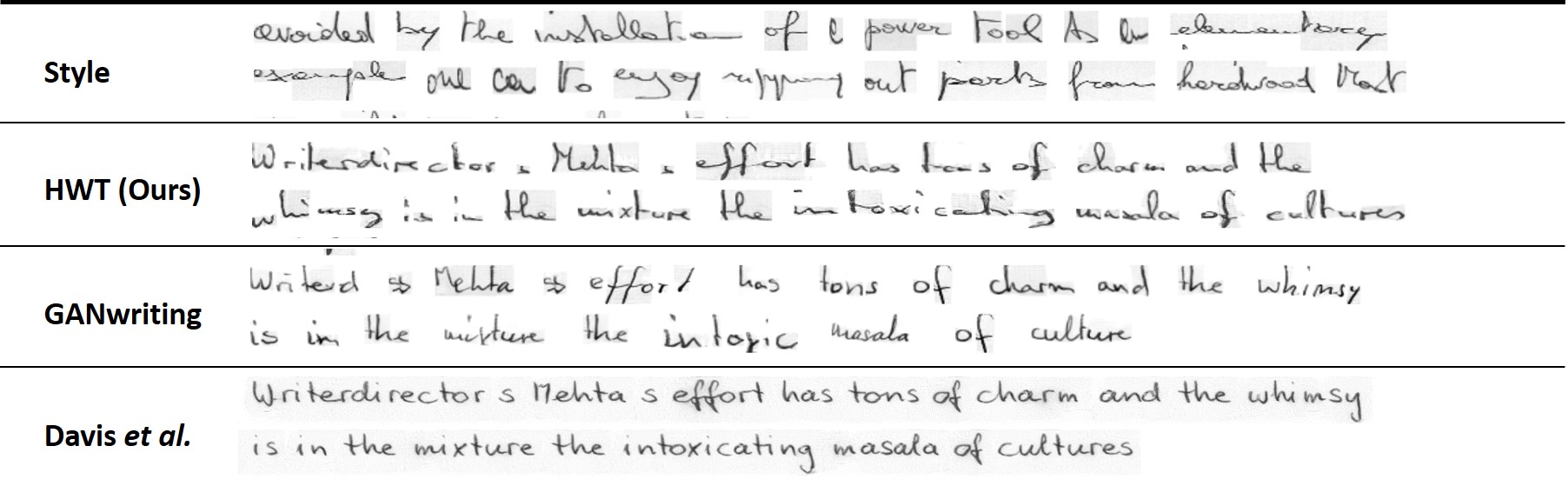}\vspace{-0.54cm}
   \label{fig:res16}
\end{center}
 \end{figure*}

\begin{figure*}[t!]
\begin{center}
\caption{Additional qualitative comparisons of our proposed HWT with GANwriting \cite{kang2020ganwriting} and Davis~\etal~\cite{davis2020text}, when generating the same text \emph{`While easier to sit through than most of Jaglom s selfconscious and gratingly irritating films it s'}.}
   \includegraphics[width=.95\textwidth]{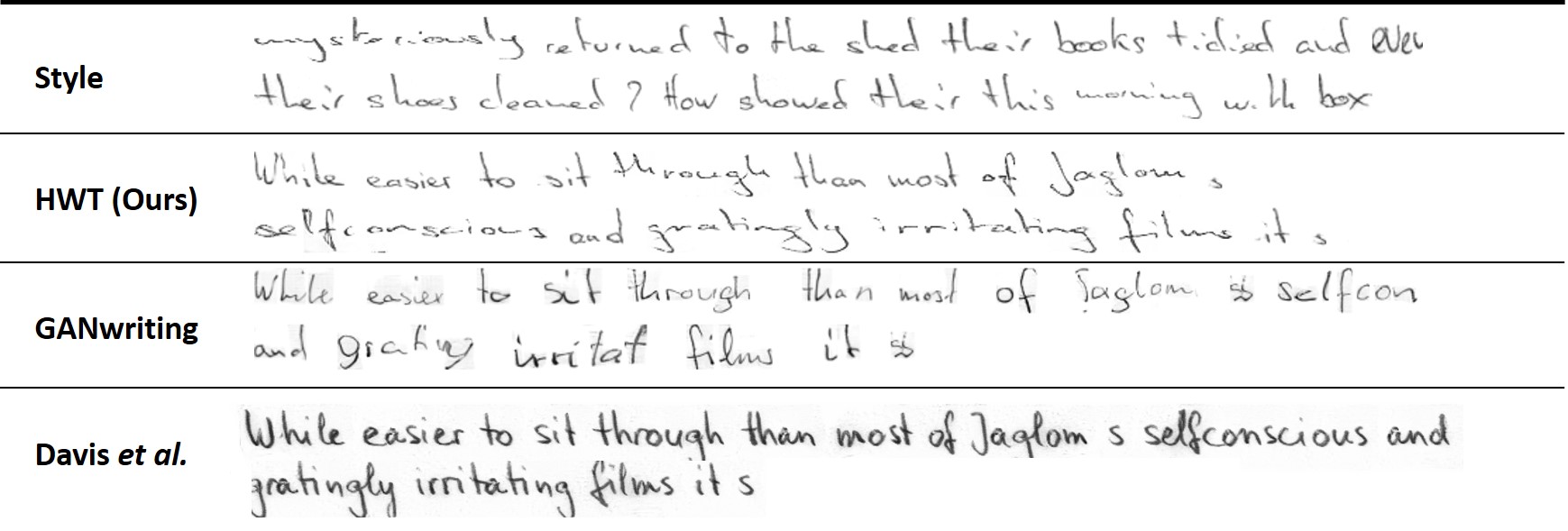}\vspace{-0.54cm}
   \label{fig:res17}
\end{center}
 \end{figure*}

\begin{figure*}[t!]
\begin{center}
\caption{Additional qualitative comparisons of our proposed HWT with GANwriting \cite{kang2020ganwriting} and Davis~\etal~\cite{davis2020text}, when generating the same text \emph{`The connected stories of Breitbart and Hanussen are actually fascinating but the filmmaking in Invincible is such that the'}. }
   \includegraphics[width=.95\textwidth]{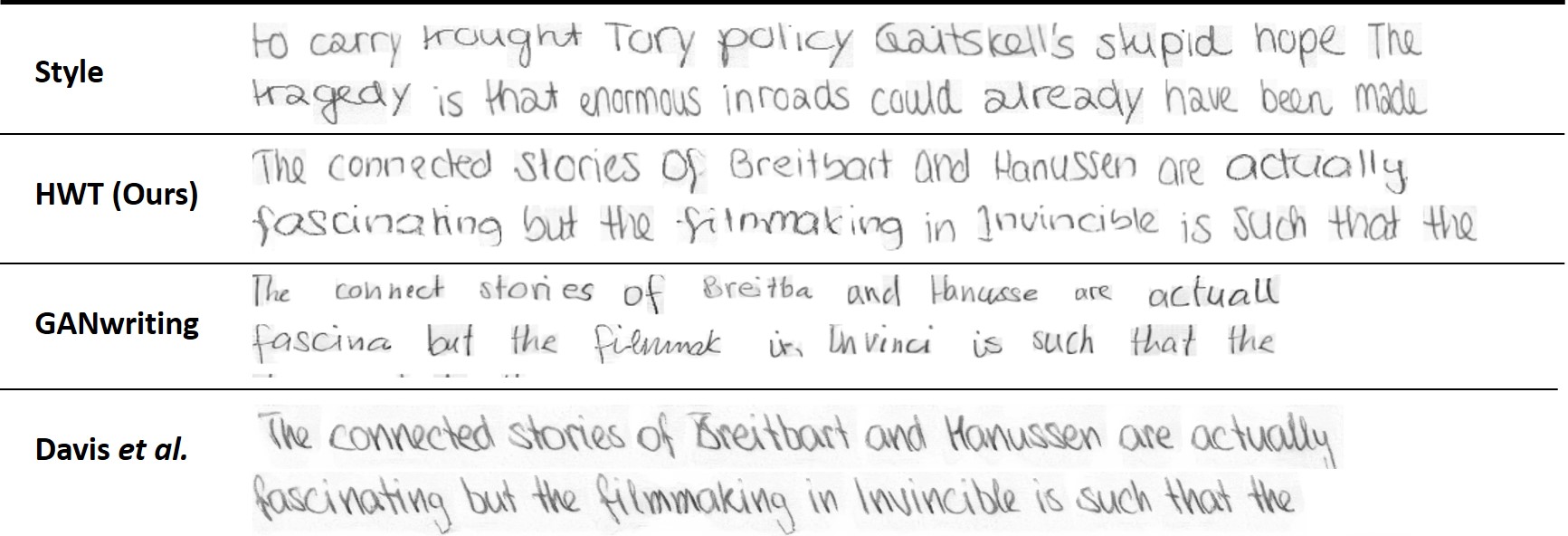}\vspace{-0.54cm}
   \label{fig:res18}
\end{center}
 \end{figure*}

Figs.~\ref{fig:res1}-\ref{fig:res18} show qualitative comparison between our proposed HWT with ~\cite{kang2020ganwriting,davis2020text} for styled handwritten text generation. Note that we use the same textual content for all the examples figures for all the three methods to ensure a fair comparison. The first row in each figure presents the different writers example style images. The rest of the rows correspond to our HWT and~\cite{kang2020ganwriting,davis2020text} respectively. The qualitative results suggest that our method is promising at imitating character-level patterns, while the other two methods struggle to retain character-specific details. The success of the other two methods is limited to capturing only the global patterns (\eg, slant, ink widths). In some cases, these methods even struggle to capture global styles. In Fig.~\ref{fig:res3}, Fig.~\ref{fig:res13} and Fig.~\ref{fig:res15}, Davis~\etal~\cite{davis2020text} suffer to capture the slant. Whereas, in Fig.~\ref{fig:res13} and Fig.~\ref{fig:res17}, the ink width of the images generated by this method is not consistent with the style examples. On the other hand, since GANwriting~\cite{kang2020ganwriting} is limited to a fixed length query words, it is unable to complete few words that exceed the limit. 

\begin{figure*}[t!]
\begin{center}
\caption{Reconstruction results using the proposed HWT in comparison to GANwriting \cite{kang2020ganwriting} and Davis~\etal~\cite{davis2020text}. We use the same text as in the style examples to generate handwritten images.}
   \includegraphics[width=1\textwidth]{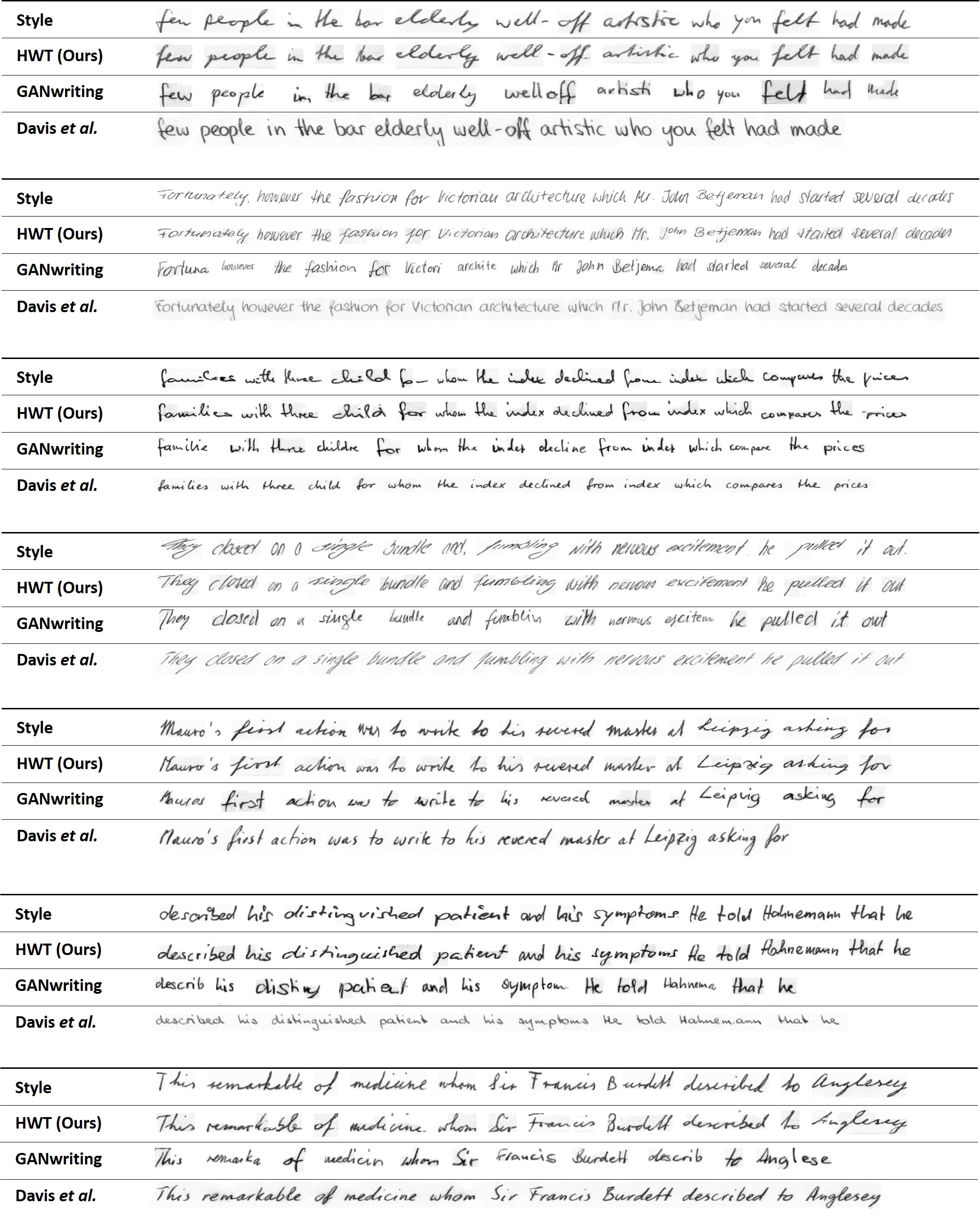}\vspace{-0.54cm}
   \label{fig:recons1}
\end{center}
 \end{figure*}

\begin{figure*}[t!]
\begin{center}
\caption{Reconstruction results using the proposed HWT in comparison to GANwriting \cite{kang2020ganwriting} and Davis~\etal~\cite{davis2020text}.}
   \includegraphics[width=1\textwidth]{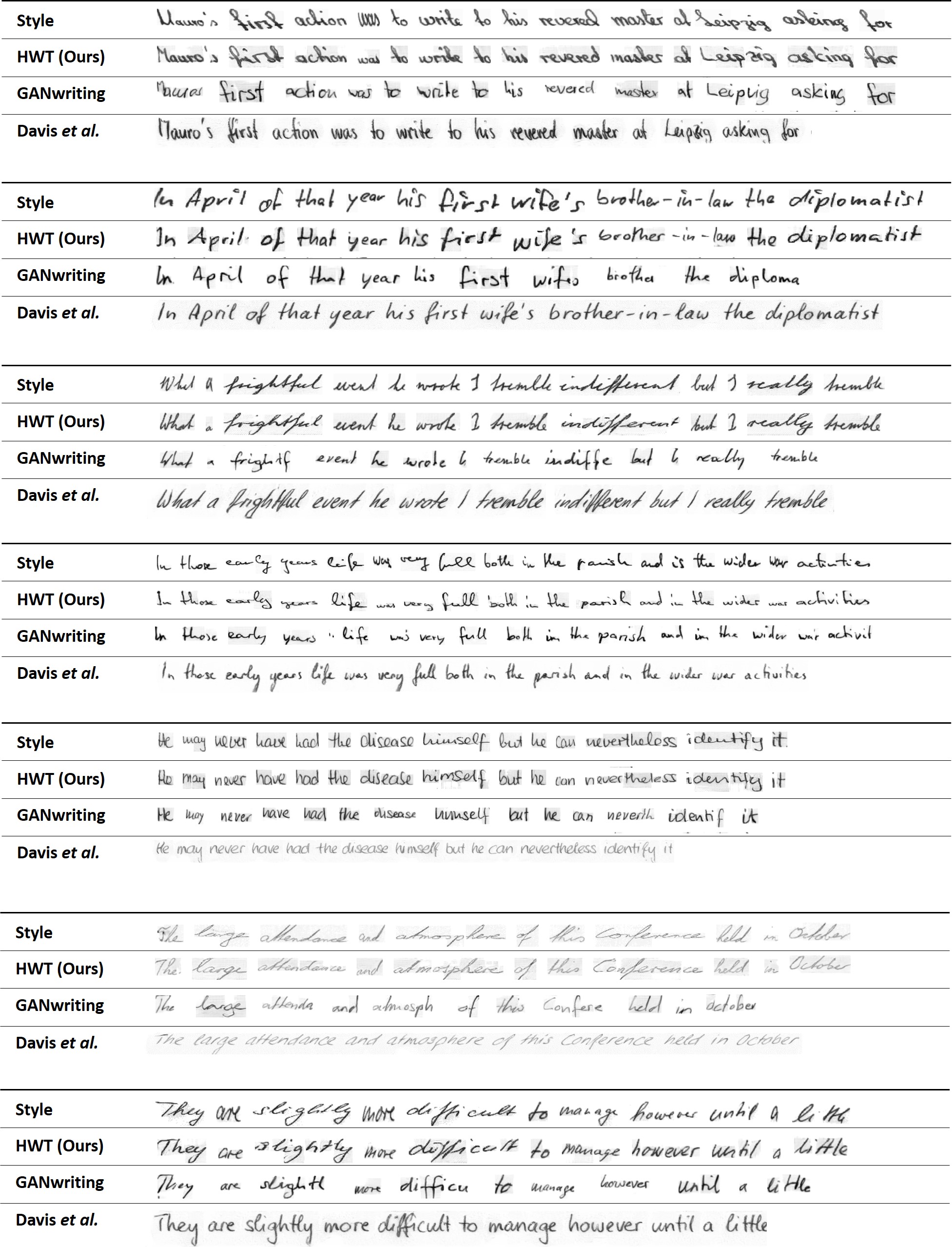}\vspace{-0.54cm}
   \label{fig:recons2}
\end{center}
 \end{figure*}

\begin{figure*}[t!]
\begin{center}
\caption{Handwritten text image generation of arbitrarily long words. We generate the 21-letter word `\wrd{Incomprehensibilities}' in three different styles and compare the results with Davis~\etal~\cite{davis2020text}.}
   \includegraphics[width=0.8\textwidth]{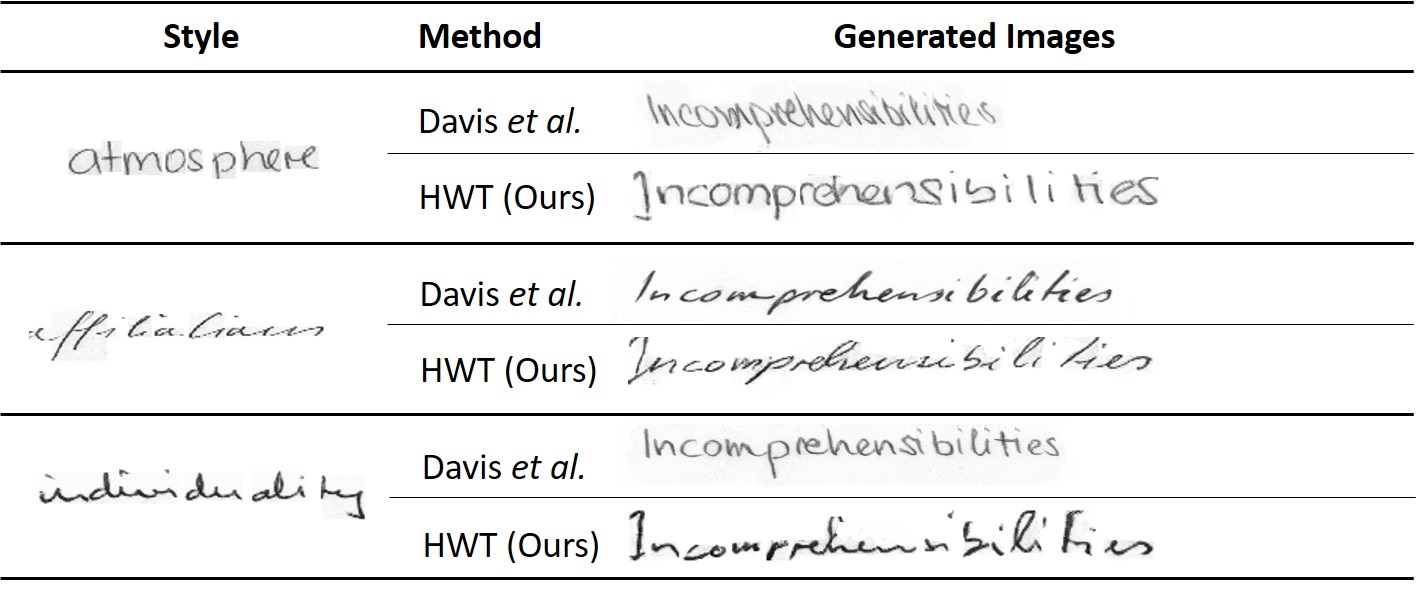}\vspace{-0.54cm}
   \label{fig:long1}
\end{center}
 \end{figure*}

\begin{figure*}[t!]
\begin{center}
\caption{Handwritten text image generation of arbitrarily long words. We generate the 30-letter word `\wrd{Pseudopseudohypoparathyroidism}' in three different styles and compare the results with Davis~\etal~\cite{davis2020text}.}
   \includegraphics[width=0.8\textwidth]{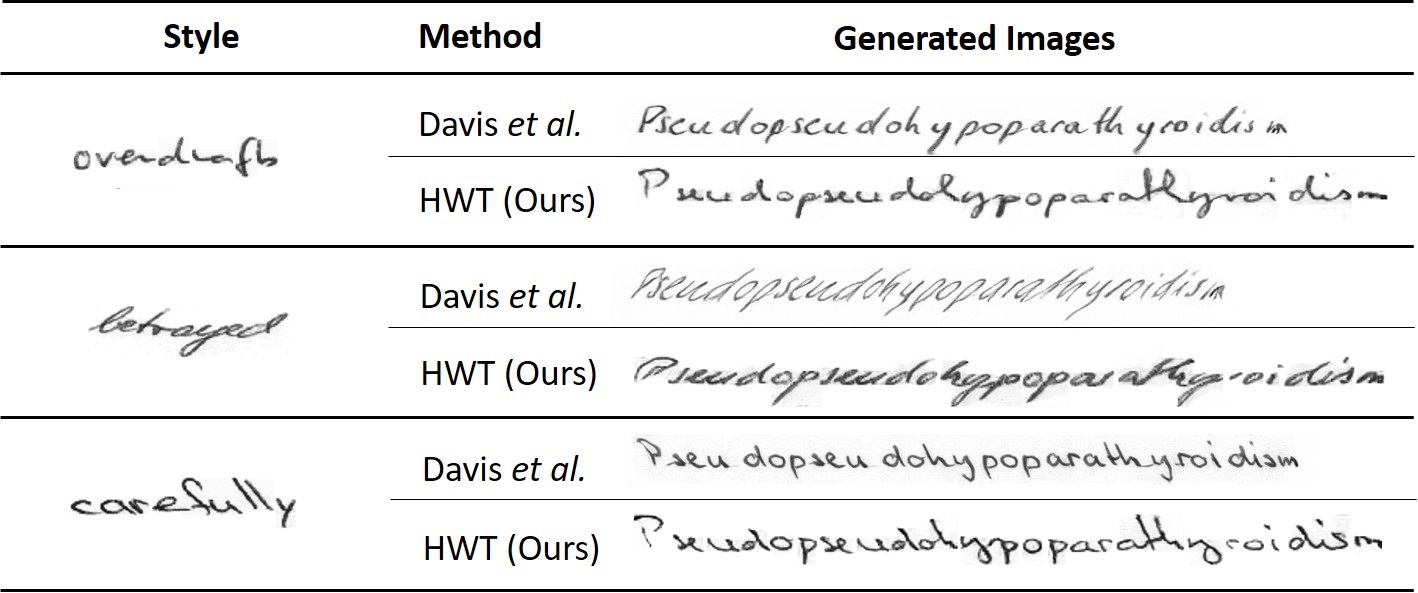}\vspace{-0.54cm}
    \label{fig:long2}
\end{center}
 \end{figure*}

\begin{figure*}[t!]
\begin{center}
\caption{Handwritten text image generation of arbitrarily long words. We generate the 28-letter word `\wrd{Antidisestablishmentarianism}' in three different styles and compare the results with Davis~\etal~\cite{davis2020text}.}
   \includegraphics[width=0.8\textwidth]{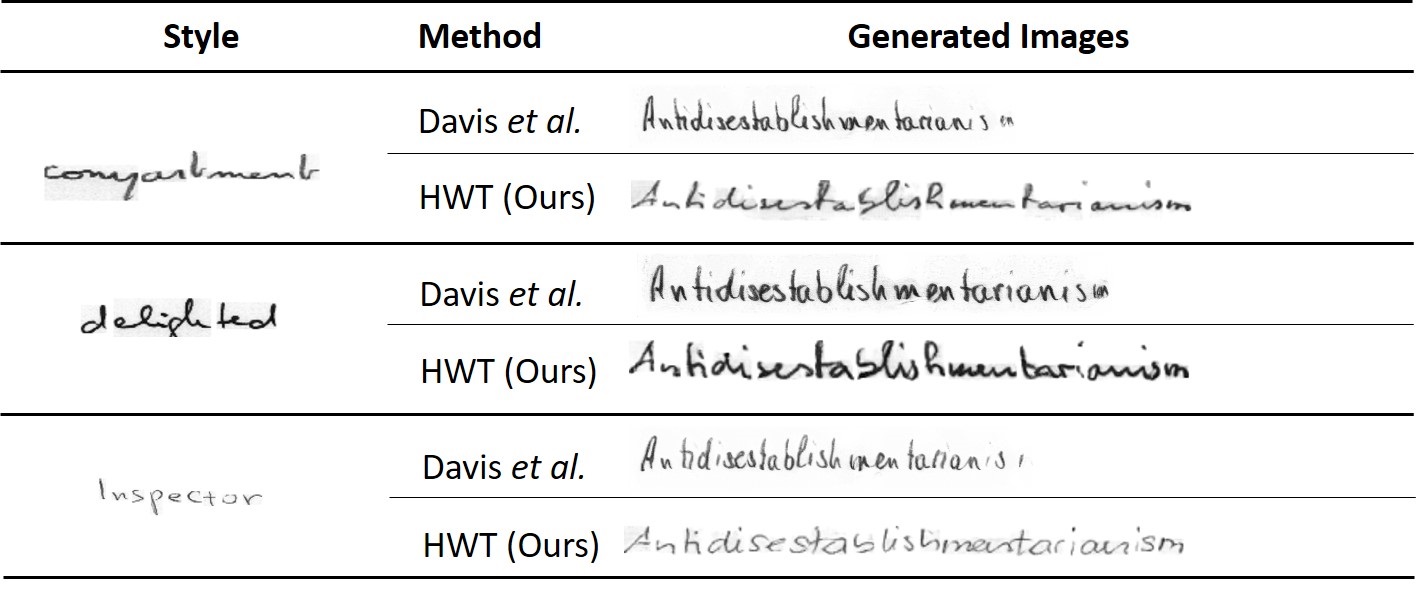}\vspace{-0.54cm}
    \label{fig:long4}
\end{center}
 \end{figure*}

Figs.~\ref{fig:recons1}-\ref{fig:recons2} show qualitative results using the same text as in the style examples to compare our proposed HWT with ~\cite{kang2020ganwriting,davis2020text}. 
Figs.~\ref{fig:long1}-\ref{fig:long4} show examples, where we aim to generate arbitrarily long words. The results show that our model is capable of consistently imitating the styles of the given style example, even for arbitrarily long words. Note that GANwriting~\cite{kang2020ganwriting} struggles to generate long words.


\section{Latent Space Interpolations}
\label{sec:interpolation}

\begin{figure*}[t!]
\begin{center}
\caption{Latent space interpolations between calligraphic styles on the IAM dataset. The first and last image in each column correspond to writing styles of two different writers. Total we have shown five sets of interpolation results. We observe how the generated images seamlessly adjust from one style to another. This result shows that our model can generalize in the latent space rather than memorizing any trivial writing patterns.}
   \includegraphics[width=1\textwidth]{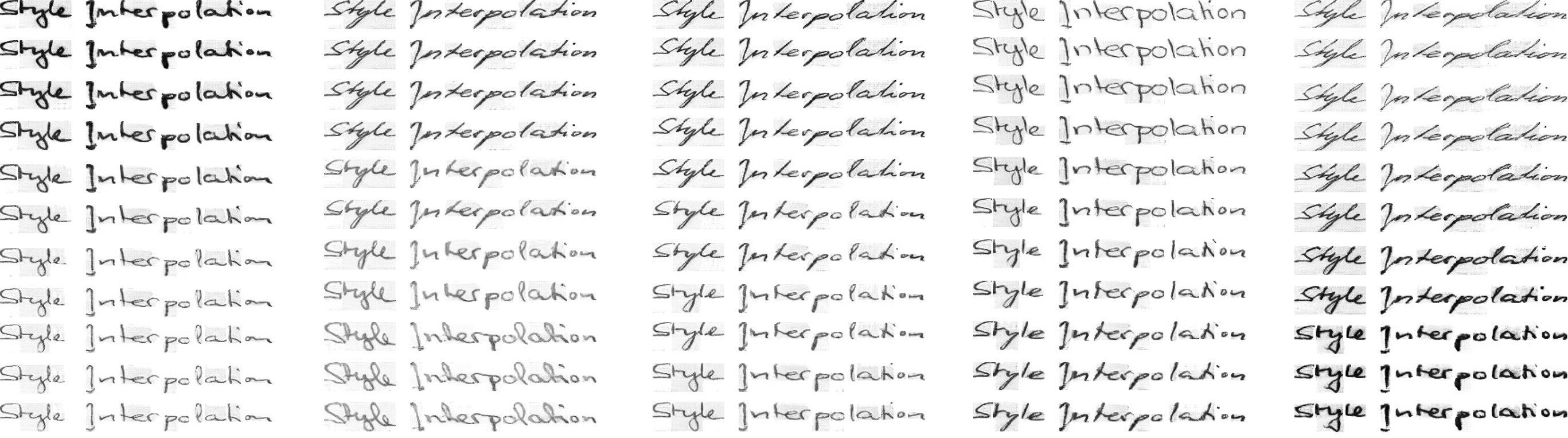}\vspace{-0.54cm}
    \label{fig:interpolation}
\end{center}
 \end{figure*}

Fig.~\ref{fig:interpolation} shows interpolations between two different calligraphic styles on the IAM dataset. To interpolate by $\lambda$ between two writers $A$ and $B$, we compute the weighted average $Z_{AB} = \lambda Z_A + (1 - \lambda )Z_B$, while keeping the textual contents fixed. Here, $Z_A$ and $Z_B$ are the style feature sequences obtained from encoder $T_{\mathcal{E}}$. It is worth  mentioning that our models produce images seamlessly by adjusting from one style to other, which indicates that our model generalizes in the latent space rather than memorizing any trivial writing patterns. 

\section{Additional Ablation Results}
\label{sec:ablation}

\begin{figure*}[t!]
\begin{center}
\caption{Additional qualitative ablation of integrating transformer encoder (Enc), transformer decoder (Dec) and cycle loss (CL) to the baseline (Base) on the IAM dataset. We show the effect of each component when generating six different words `\wrd{especially}', `\wrd{ethereal}', `\wrd{emotional}', `\wrd{standard}',`\wrd{resorts}', and `\wrd{under}'.}
   \includegraphics[width=0.9\textwidth]{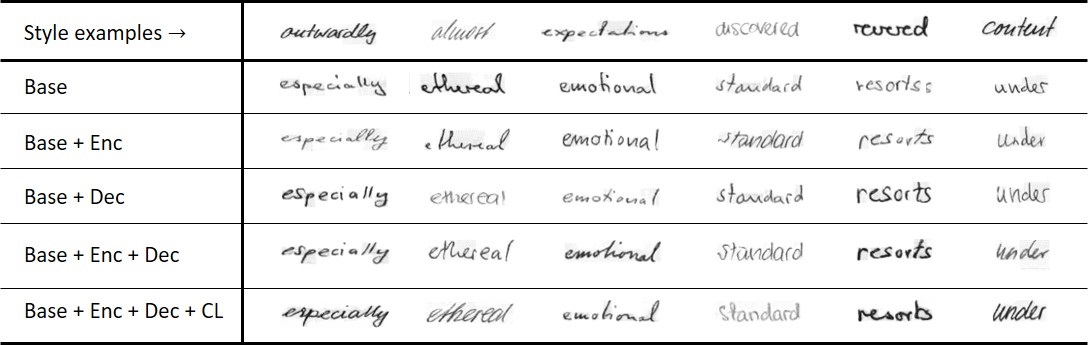}\vspace{-0.54cm}
   \label{fig:ablation1}
\end{center}
 \end{figure*}

\begin{figure*}[t!]
\begin{center}
\caption{Additional qualitative comparisons between word and character-level conditioning on IAM dataset. We show the comparison between word and character-level conditioning when generating six different words `\wrd{engaging}', `\wrd{actually}', `\wrd{movie}', `\wrd{rhythms}',`\wrd{what}', and `\wrd{evocative}'.}
   \includegraphics[width=0.9\textwidth]{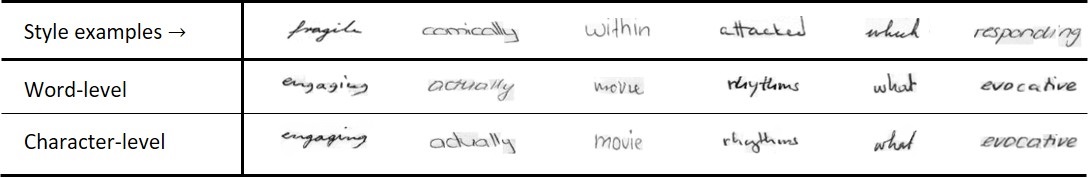}\vspace{-0.54cm}
   \label{fig:ablation2}
\end{center}
 \end{figure*}

Fig.~\ref{fig:ablation1} presents additional qualitative results that show the impact of integrating transformer encoder  (Enc),  transformer  decoder  (Dec)  and  cycle  loss (CL) to the baseline (Base). Fig.~\ref{fig:ablation2} shows additional qualitative comparisons between word-level and character-level conditioning.

\begin{figure*}[t!]
\begin{center}
\caption{Additional visualization results of encoder-decoder attention maps at the last layer of the transformer decoder. The attention maps are computed for four different query words: `\wrd{laughs}', `\wrd{because}', `\wrd{inside}', and `\wrd{fashion}'. Heat maps corresponding to all characters (including repetitions, as the letter `\wrd{i}' appears twice in `\wrd{inside}') of these query words are shown in the figure. }

   \includegraphics[width=1\textwidth]{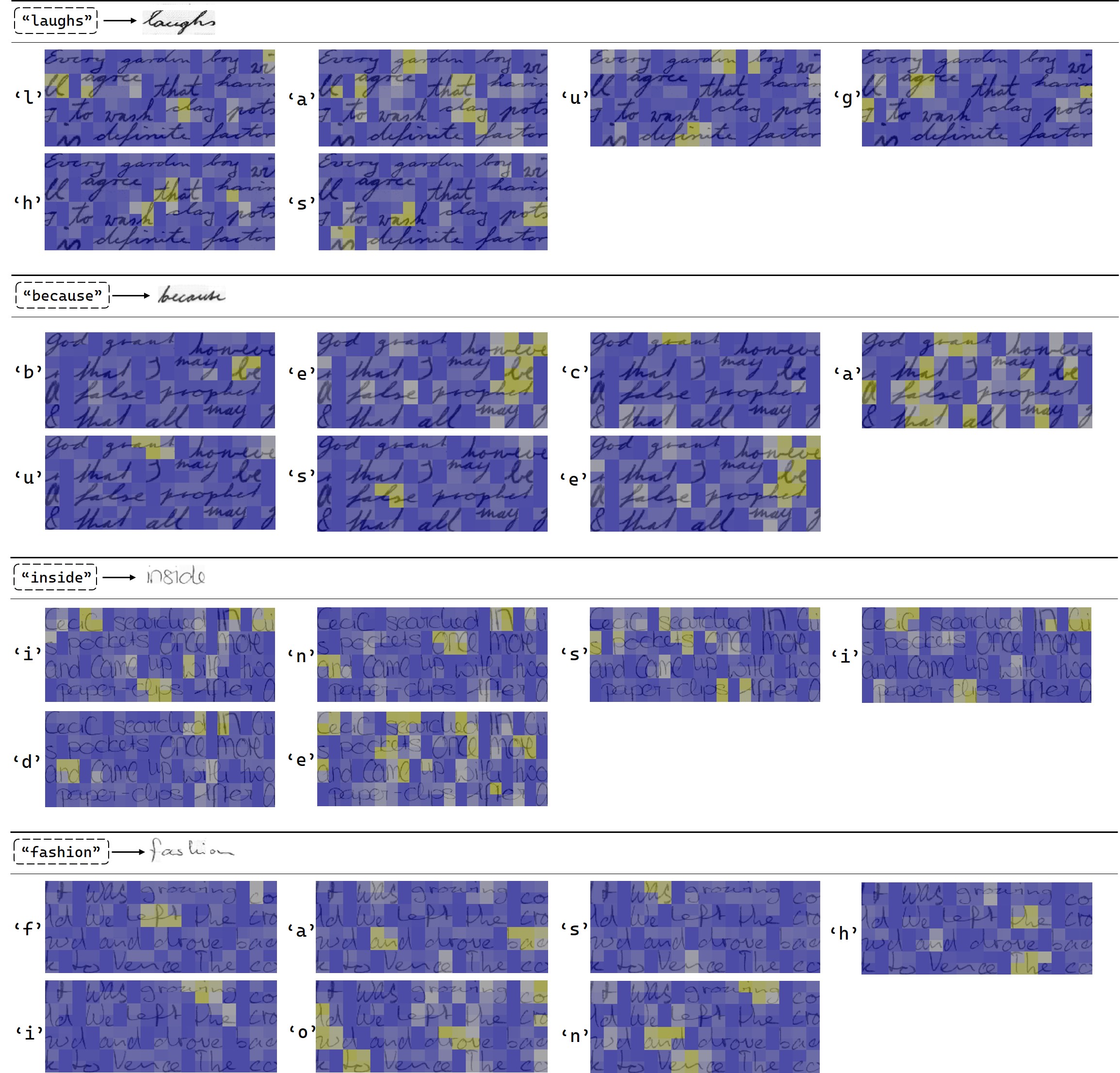}\vspace{-0.54cm}
   \label{fig:attention}
\end{center}

 \end{figure*}

\end{document}